\documentclass[10pt,journal,compsoc]{IEEEtran}

\IEEEoverridecommandlockouts
\usepackage{cite}
\usepackage{amsmath,amssymb,amsfonts}
\usepackage{algorithm}
\usepackage[noend]{algpseudocode}
\usepackage{graphicx}
\usepackage{balance}
\usepackage{xcolor}
\def\BibTeX{{\rm B\kern-.05em{\sc i\kern-.025em b}\kern-.08em
    T\kern-.1667em\lower.7ex\hbox{E}\kern-.125emX}}

\usepackage{array}
\usepackage{textcomp}
\usepackage{stfloats}
\usepackage{url}
\usepackage{verbatim}
\usepackage[utf8]{inputenc}
\usepackage{booktabs}

\newcommand{\model}{\textsf{CHASe}}
\newcommand{\FAMS}{\textsf{FAmS}}
\newcommand{\fram}{FAL}
\newtheorem{lemma}{\bf Lemma}
\newenvironment{proof}{{\bf Proof.~}\it}{\hfill $\square$\par\bigskip}
\usepackage{caption} 
\usepackage{pifont} 
\usepackage{amssymb}
\usepackage{multirow} 
\usepackage[normalem]{ulem}
\useunder{\uline}{\ul}{}
\usepackage{subfigure}
\usepackage{float}
\usepackage[switch]{lineno}

\usepackage[shortlabels]{enumitem}
\let\temp\rmdefault
\usepackage{mathpazo}
\let\rmdefault\temp
\usepackage{dsfont}
\usepackage{siunitx}
\usepackage[colorinlistoftodos,prependcaption,textsize=tiny]{todonotes}

\usepackage[normalem]{ulem}
\useunder{\uline}{\ul}{}

\hypersetup{
    colorlinks=false,
    pdfborder={0 0 0},
}
\usepackage[commandnameprefix=ifneeded,todonotes={textsize=scriptsize},markup=underlined]{changes}
\definechangesauthor[name={R1}, color=black]{R1}
\definechangesauthor[name={R2}, color=black]{R2}
\definechangesauthor[name={R3}, color=black]{R3}
\setcommentmarkup{%
    \IfIsColored%
        {\colorlet{Changes@todocolor}{authorcolor}}%
        {\colorlet{Changes@todocolor}{black}}%
    \todo[color=Changes@todocolor!10, bordercolor=Changes@todocolor, linecolor=Changes@todocolor!70, nolist]{%
        \textbf{#3.} #1%
    }%
}
\usepackage[capitalize]{cleveref}

\begin{document}
\title{
\model{}: Client Heterogeneity-Aware Data Selection for Effective Federated Active Learning}

\author{
Jun Zhang, Jue Wang, Huan Li,~\IEEEmembership{Member,~IEEE}, Zhongle Xie,~\IEEEmembership{Member,~IEEE}, Ke Chen, Lidan Shou
\IEEEcompsocitemizethanks{\IEEEcompsocthanksitem 
This work is supported by the Major Research Program of Zhejiang Provincial Natural Science Foundation (No.~LD24F020015),
the Pioneer R\&D Program of Zhejiang (No.~2024C01021), ``Leading Talent of Technological Innovation Program'' of Zhejiang Province (No.~2023R5214), NSFC Grant No.~62402420, and the Major Project of National Social Science Foundation (No.~24ZDA092). (Corresponding authors: Huan Li; Lidan Shou.) \\
J.~Zhang, J.~Wang, H.~Li, Z.~Xie, K.~Chen and L.~Shou are with the State Key Laboratory of Blockchain and Data Security, Zhejiang University, China, and also with Hangzhou High-Tech Zone (Binjiang) Institute of Blockchain and Data Security. 
E-mail: \{zj.cs, zjuwangjue, lihuan.cs,	xiezl, chenk, should\}@zju.edu.cn
}
}



\IEEEtitleabstractindextext{
\begin{abstract}
Active learning (AL) reduces human annotation costs for machine learning systems by strategically selecting the most informative unlabeled data for annotation, but performing it individually may still be insufficient due to restricted data diversity and annotation budget.
Federated Active Learning (FAL) addresses this by facilitating collaborative data selection and model training, while preserving the confidentiality of raw data samples. 
Yet, existing FAL methods fail to account for the heterogeneity of data distribution across clients and the associated fluctuations in global and local model parameters, adversely affecting model accuracy.
To overcome these challenges, we propose \model{} (Client Heterogeneity-Aware Data Selection), specifically designed for FAL. \model{} focuses on identifying those unlabeled samples with high epistemic variations (EVs), which notably oscillate around the decision boundaries during training.
To achieve both effectiveness and efficiency, \model{} encompasses techniques for 1) tracking EVs by analyzing inference inconsistencies across training epochs, 2) calibrating decision boundaries of inaccurate models with a new alignment loss, and 3) enhancing data selection efficiency via a data freeze and awaken mechanism with subset sampling.
Experiments show that \model{} surpasses various established baselines in terms of effectiveness and efficiency, validated across diverse datasets, model complexities, and heterogeneous federation settings.
\end{abstract}

}

\maketitle

\IEEEdisplaynontitleabstractindextext

%
\IEEEpeerreviewmaketitle

\section{Introduction}
\label{sec:intro}

\IEEEPARstart{I}{nadequate} labeled data is a common challenge faced by machine learning systems, primarily due to the high cost of annotations \cite{9835163,9458763}.
To address this issue, researchers have proposed Active Learning (AL) \cite{cohn1996active}, which aims to learn an accurate model with optimal annotation cost. AL achieves this goal by wisely identifying the most informative unlabeled data and querying their labels from human experts.

However, individual data owners performing AL often face constraints due to limited human resources and the isolated nature of data samples, which can hinder achieving satisfactory model accuracy. 
The ideal solution involves collaborative efforts from multiple data owners in annotation and training \cite{ijcai2021-354,8057034}. However, such collaboration raises concerns regarding data privacy, as direct sharing of raw data for annotation and training may not always be feasible or appropriate.




\begin{figure}[t]
    \centering
    \includegraphics[width=1\linewidth]{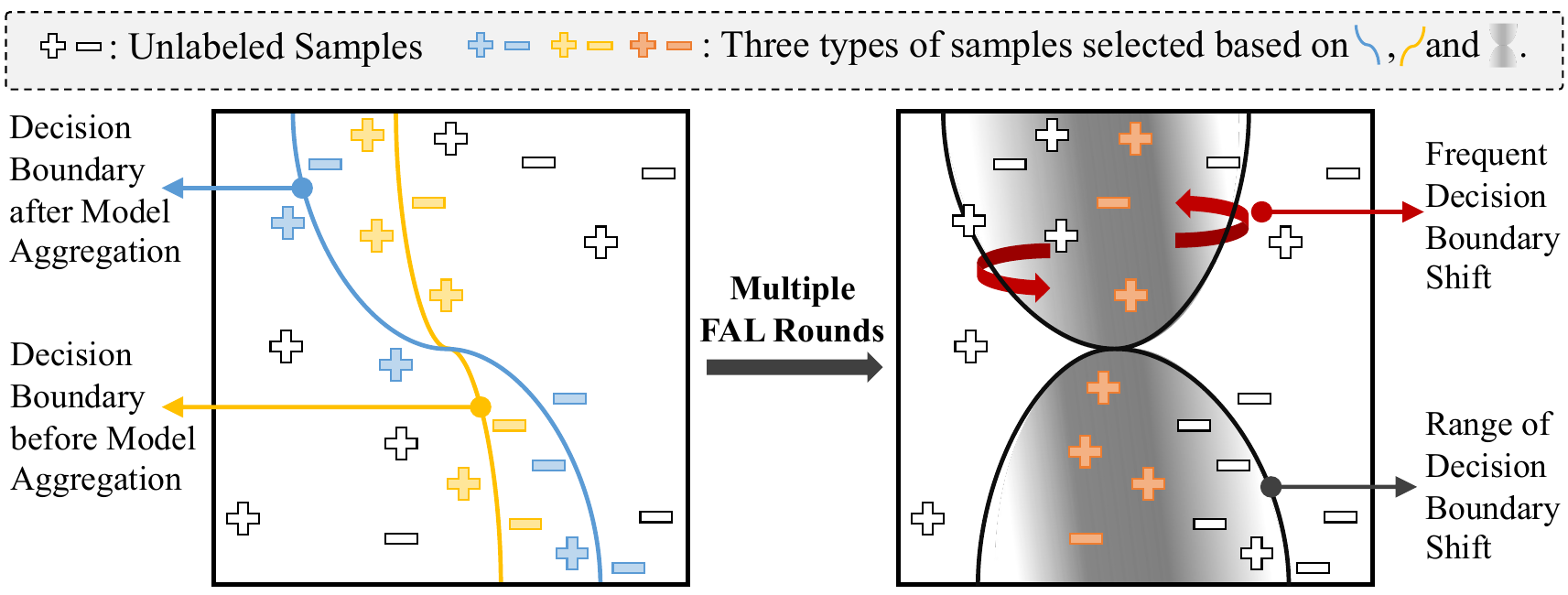}
    \caption{Visualization of samples' epistemic variation in \fram{}.}
    \label{fig:intro}
\end{figure}

Federated Learning (FL) \cite{mcmahan2017communication, li2018federated,chen2019communication,9835537,DM-PFL} has been proposed for conducting machine learning in a decentralized manner by communicating the parameters instead of raw data.
The main idea is to obtain a \emph{global} shared model by aggregating models trained locally on the distributed clients \cite{9458704}.
Driven by the potential benefits of leveraging other clients' data and annotation capabilities, the concept of \emph{Federated Active Learning} (\fram{}) is emerging.
In \fram{}, each client performs AL to annotate its data, trains a local model with newly labeled data, and sends its model update to a central server to obtain a consensus model. 
{In traditional FAL, data selection spans the entire FL training cycle~\text{\cite{cao2022knowledgeaware,kim2023rethinking,jin2022federated}}. Here, we focus on scenarios where client participation in communication rounds is intermittent or uncertain, such as mobile devices with unstable network connections~\text{\cite{saha2022colrel}} or tasks like next-word prediction~\text{\cite{mcmahan2017communication}}, where clients participate only in short, burst-like interactions. To address this, we enable data selection to occur immediately when a client joins a communication round, maximizing the collection of valuable data, even during brief participation. The key distinction of our FAL is its ability to perform high-informative sampling based on knowledge gained from each communication round—specifically, the interaction between local and global models. Our setup represents a specific case of traditional FAL with continual initialization, where FL is performed by leveraging the previously learned model.
In~\text{\cref{exp:continual_fal}}, our proposal proves robust to an alternative FAL setup with random initialization.
}
Our study in \cref{sssec:necessity_fal} reveals that our \fram{} substantially surpasses traditional, isolated AL approaches in accuracy.

Current \fram{} studies focus on designing effective acquisition strategies, e.g.,~entropy-based \cite{luo2013latent}, Core-set \cite{sener2018active}, and LL4AL \cite{yoo2019learning}, to pick out data residing at the model's decision boundary. An example is depicted in~\cref{fig:intro} (left).
This process involves choosing samples at each \fram{} round, based on either the local model's boundary before aggregation (see the yellow curve and markers in \cref{fig:intro}) \cite{jia2019active} or the post-aggregation global model (blue curve and markers) \cite{jin2022federated}. 

However, in practical scenarios with Non-Independent-and-Identically-Distributed (Non-IID) data across varied devices \cite{10184650}, these methods often do not perform optimally. Our empirical analysis in~\cref{ssec:exp_overall} indicates that they might even be less effective than random data selection. The Non-IID nature of data causes notable differences between the parameters of global and local models, resulting in a clear disparity in the samples chosen based on their respective decision boundaries (yellow and blue markers in~\cref{fig:intro} show little overlap). It becomes a challenge for clients to discern which set of samples is more informative due to the unstable and varying boundaries before and after each round's aggregation.

Some methods attempt to address these issues by analyzing sample diversity \text{\cite{cao2022knowledgeaware}} and majority \text{\cite{kim2023rethinking}} at the decision boundaries of both models, before and after aggregation, potentially combining selections from both yellow and blue markers.
However, these methods fail to consider the inherent inaccuracy of the decision boundaries at any instant time, rendering the combined set of blue and yellow samples less effective. 
Moreover, as the \fram{} progresses through multiple rounds, the continuous evolution of client data distributions leads to significant shifts in decision boundaries (represented by a grey area in~\cref{fig:intro}).
This situation is further complicated by the behavioral heterogeneity of clients, i.e.,~variations in annotation rounds and volumes, as discussed in~\cref{ssec:exp_overall}).

In an ideal setting, the decision boundary of a well-trained model should remain stable.
However, in \fram{}, ongoing shifts in this boundary indicate prolonged model convergence, suggesting the model has not fully captured the underlying data distribution.
Furthermore, samples near these unstable boundaries often highlight regions of high model uncertainty, reflecting incomplete or conflicting knowledge. 
Given FAL’s annotation cost constraints, selecting all samples during boundary shifts is impractical. Thus, prioritizing those frequently near unstable boundaries is crucial. This strategy accelerates convergence and boosts global model performance under Non-IID conditions in \fram{} (see\text{~\cref{ssec:exp_overall}}).
Closer examination reveals that, as FAL progresses, there is a noticeable shift in the inference results for certain unlabeled samples alongside the boundary's shift. We refer to this as samples' \textbf{epistemic variation} (EV).
As depicted in\text{~\cref{fig:intro}}, during multiple FAL rounds, samples marked in orange demonstrate high EV due to their frequent interaction with the decision boundary (with darker shades indicating higher frequency). Therefore, focusing on these samples for human annotation is beneficial.
In response, we introduce a specialized method named \model{} ({C}lient {H}eterogeneity-{A}ware Data {Se}lection), designed to boost the effectiveness and efficiency of \fram{}.
Essentially, \model{} targets unlabeled samples with high EVs --- those fluctuating near the decision boundaries --- for human annotation. 
\model{} aims to optimize the impact of \fram{} within a limited annotation budget and it encompasses a complete set of techniques to achieve this goal.

First of all, we propose a method to quantify the EV for each candidate unlabeled sample.
This is achieved by tracking the number of times a sample yields inconsistent inference outcomes across two consecutive epochs. Essentially, EV serves as an indicator of how challenging it is for a sample to be consistently classified by the local model. 
As demonstrated in \cref{ssec:ev_relevance}, there is a significant correlation between the measured EVs and the ultimate accuracy of the model.

Quantifying EVs effectively is however challenging in a Non-IID setting, primarily due to inaccuracies in model inference. This issue arises from two factors: first, the heterogeneity of local models, each tailored to its specific local data distribution, and second, the construction of the global model from such heterogeneous local models. To counter this, we introduce a technique to calibrate the decision boundaries of local models, thereby enhancing the accuracy of both local models and their collectively formed global model. Our approach involves designing a novel alignment term that serves two purposes: (1) it ensures that `\emph{simple}' samples, featured by small EVs, maintain consistent representations across consecutive local training epochs, and (2) it aligns the representation of `\emph{hard}' samples, marked by large EVs, within the global model --- the rationale behind using the global model for aligning hard samples is its comprehensive knowledge derived from the collective input of multiple clients.

Quantifying EVs for samples also involves exhaustive inferences at each training round.
Thus, we propose an efficiency optimization technique for \model{}, in which a data freeze and awaken mechanism with subset sampling (\FAMS{}) is proposed.
Specifically, samples exhibiting zero EVs, deemed less informative, are `frozen' in the initial stages to cut down on inference workload. Additionally, subset sampling is utilized to further diminish the pool of samples under consideration, streamlining the process.

We compare multiple AL methods under various federation settings, validating the effectiveness and efficiency of our proposals on image and text datasets. The empirical evidence shows that \model{} achieves superior performance across a range of tasks and scenarios.

Our main contributions are summarized as follows.
\begin{itemize}[leftmargin=*]
\item We identify the sample's epistemic variation (EV) phenomenon in federated active learning (\fram{}) with client heterogeneity, and utilize it to tackle the practical problem of \fram{} being limited by annotation budget (see \cref{sec:problem}).

\item We propose a novel data selection method \model{}, which leverages EVs to guide selecting highly informative samples in \fram{} (see \cref{subsec:CEV}).
\model{} also utilizes EVs to design an alignment loss to calibrate models' decision boundaries during \fram{}, so as to enhance the effectiveness of data selection in Non-IID settings (see \cref{subsec:CDB}).

\item We design a data freeze and awaken mechanism with subset sampling (\FAMS{}) to optimize the efficiency of \model{}, making it cost-effective in \fram{} scenarios (see \cref{subsec:EO}).

\item We conduct extensive experiments on benchmark image and text datasets, verifying the proposals' effectiveness compared to multiple baselines and its generalizability in different datasets and federation scenarios (see \cref{sec:experiment}).

\end{itemize}
In addition, \cref{sec:related} reviews the inspiring works and \cref{sec:conclusion} concludes the paper with future directions.
\section{Problem Definition}
\label{sec:problem}

\cref{ssec:AL-fram} and \cref{ssec:FAL-fram} present the frameworks of active learning and federated active learning, respectively.
The frequently used notation in this paper is provided in 
supplementary material \text{\cite{github-chase}}.

\subsection{Framework of Active Learning (AL)}\label{ssec:AL-fram}

AL~\cite{cohn1996active} aims to obtain a high-quality model with a relatively small labeled set by iteratively selecting the most informative samples for annotation from the unlabeled data pool.
AL assumes that the user initially has a labeled set with $C$ classes $\mathcal{D}^{L}=\{\boldsymbol{x}^{i},y^{i}\}_{i=1}^{N^{L}}$ and an unlabeled set $\mathcal{D}^{U}=\{\boldsymbol{x}^{i}\}_{i=1}^{N^{U}}$, where $\boldsymbol{x}^{i}$ and $y^{i}$ represents a sample and its corresponding label, respectively.
The sizes of two sets satisfy that ${N^{L}} \ll{N^{U}}$.

For the sampling round $r$, AL first trains the model $\boldsymbol{\omega}^{r}$ on current labeled set $\mathcal{D}^{L}$. Then unlabeled samples in $\mathcal{D}^{U}$ will be evaluated on an \emph{acquisition strategy} \cite{ShaoLLCYLX22,cohn1996active}.
According to the evaluation result, the set $\mathcal{D}^{S}$ of highly informative samples will be selected from $\mathcal{D}^{U}$ and annotated with the fixed budget $\mathcal{B}$.
We use the number of samples as the unit for the budget.
The labeled and unlabeled sets will then be updated, formally $\mathcal{D}^{L} \leftarrow \mathcal{D}^{L} \cup \mathcal{D}^{S}$ and $\mathcal{D}^{U} \leftarrow \mathcal{D}^{U} \setminus \mathcal{D}^{S}$. 
This process is repeated until the specified model accuracy or the maximal sample round $R$ is met.

\subsection{Framework of Federated Active Learning (\fram{})} \label{ssec:FAL-fram}

\fram{}~\cite{jia2019active,jin2022federated} extends the conventional AL to federated learning with $K$ clients and a server. Each client $k$ annotates their own unlabeled data $\mathcal{D}_{k}^{U}$, trains models locally on new labeled set $\mathcal{D}_{k}^{L}$ and learns a shared model via $R$ rounds of communications of model parameters with the server, overcoming the limitations of data range and human efforts while maintaining data privacy.

At round $r \in [1,R)$, the client $k$ trains the received global model $\boldsymbol{\omega}^{r-1}$ with $E$ epochs on the labeled set by minimizing the loss function $\ell(\cdot)$, as follows:  
\begin{equation}
\boldsymbol{\omega}_{k}^{r} \leftarrow \boldsymbol{\omega}^{r-1}-\eta \nabla_{\boldsymbol{\omega}} \ell(\boldsymbol{\omega}^{r-1};
\mathcal{D}_{k}^{L}),
\end{equation}
where $\eta$ is the learning rate. Then, the client $k$ uploads the local model $\boldsymbol{\omega}_{k}^{r}$ to the server.
The local model at any epoch $e < E$ is called a \emph{partially trained local model}.

The server aggregates all received local updates by the weighted average (e.g.,~FedAvg \cite{mcmahan2017communication}) to update the 
global model parameters \footnote{We assume the commonly used weighted average strategy.}
, as follows: 
\begin{equation}
\boldsymbol{\omega}^{r} \leftarrow \sum_{k=1}^{K} \frac{N_{k}^{L}}{N} \boldsymbol{\omega}_{k}^{r},
\end{equation}
where $N=\sum_{k=1}^{K} N_{k}^{L}$. Subsequently, these updated parameters are distributed to all the clients in the federation.

{In traditional FAL, informative sample selection, as detailed in \text{\cref{ssec:AL-fram}}, occurs after completing the entire FL training cycle with $R$ rounds. In a typical setup, the next full FL cycle may choose between random initialization and continual initialization based on the current learned model. However, under the continual initialization setting, the learned model may be affected by the heterogeneous data distributions across clients. Hence, random initialization before FL is considered a standard setting of traditional FAL~\text{\cite{cao2022knowledgeaware,kim2023rethinking,jin2022federated}}.
Our \text{\fram{}}, on the other hand, focuses on scenarios where client participation is intermittent or uncertain~\text{\cite{saha2022colrel,mcmahan2017communication}}. To address this, we enable data selection as soon as a client joins a communication round $r$, ensuring that valuable client data is collected promptly, even during brief participation. Specifically, once a client receives $\boldsymbol{\omega}^{r}$, it performs the aforementioned AL procedure locally}\footnote{{As clients may drop out each round, we perform AL after every communication to maximize annotation capacity. \text{\cref{exp:continual_fal}} shows \model{} remains effective in traditional FAL with larger sampling intervals and random model initialization prior to executing FL.}}: 
It first picks a subset $\mathcal{D}_{k}^{S}$ 
with the budget $\mathcal{B}_{k}$
from the unlabeled set via an acquisition strategy $A(\cdot)$, queries their labels from human experts, and adds them to the labeled set, formally
\begin{equation}
    \mathcal{D}_{k}^{L} \leftarrow \mathcal{D}_{k}^{L} \cup \mathcal{D}_{k}^{S}; \,\,\,\,\,\,
    \mathcal{D}_{k}^{U} \leftarrow \mathcal{D}_{k}^{U} \setminus \mathcal{D}_{k}^{S}.
\end{equation}

The above steps are repeated and the end condition is akin to that in traditional AL.
Given a fixed overall budget $\mathcal{B}$, the objective function of \fram{} is formulated as minimizing the expected local model losses:
\begin{equation} \label{eq4}
    \begin{split}
    & \min\nolimits_{(\sum_{k=1}^{K}\mathcal{B}_{k}) \leq \mathcal{B}} \sum\nolimits_{k=1}^{K} \frac{N_{k}^{L}}{N} \mathbb{E}[\ell(\boldsymbol{\omega}^{r-1}; \mathcal{D}_{k}^{L} \cup \mathcal{D}_{k}^{S})], \\
    &\mathcal{D}_{k}^{S} = \{\boldsymbol{x}^{i} \mid A(\boldsymbol{\omega}_{k}^{r};\boldsymbol{\omega}^{r};\mathcal{D}_{k}^{U};\mathcal{B}_{k})\}.
    \end{split}
\end{equation}

In light of the above objective function, we establish the following lemma.

\begin{lemma}[NP-Hardness in \fram{}]\label{lemma_1}
In a scenario where each client $k$ has a non-empty unlabeled dataset ($D_k^U \neq \emptyset$) and operates under a fixed overall budget $\mathcal{B}$, the objective function for \fram{} is NP-hard.
\end{lemma}

\begin{proof}\label{proof_1}
If the unlabeled set of each client is empty, \fram{} degenerates to the conventional FL.
However, with non-empty unlabeled datasets, the problem transforms into a 0/1 knapsack problem \cite{karp2010reducibility}.
Here, the annotation budget per client $\mathcal{B}_{k}$ represents the \emph{capacity} and the empirical risk $\mathbb{E}[\cdot]$ serves as the \emph{utility function}.
The total number of possible subsets is of $\mathcal{O}(2^n)$ with $n=N_{k}^{U} = |\mathcal{D}_{k}^{U}|$ and the minimum complexity for verifying a subset is of $\mathcal{O}(n)$.
Hence, the computational complexity of the framework is $\mathcal{O}(n \times 2^{n})$.
Being equivalent to 0/1 knapsack problem, \fram{} with budget $\mathcal{B}$ is NP-hard.
\end{proof}


Given the NP-hardness, traditional AL~\cite{luo2013latent,sener2018active,yoo2019learning} converts the acquisition strategy into a process of greedily choosing the subset of samples with the currently highest significance value derived from the trained model.
Those approaches work well in FL under IID settings, which assume
the expectation of loss for local and global models is the same, i.e.,~$\mathbb{E}[\ell(\boldsymbol{\omega}^{r-1};\mathcal{D}_{k}^{L} \cup \mathcal{D}_{k}^{S})]=\mathbb{E}[\ell(\boldsymbol{\omega}^{r-1}; \mathcal{D}^{L} \cup \mathcal{D}^{S})]$.
However, this IID assumption falls short in heterogeneous \fram{} scenarios, where a low loss in local models does not necessarily imply an effective global model, and vice versa.
Consequently, relying solely on a local model (ante-aggregation)\text{~\cite{jia2019active}}, a global model (post-aggregation)\text{~\cite{jin2022federated}}, or merely combining these two instantaneous models\text{~\cite{cao2022knowledgeaware,kim2023rethinking}}, may not yield the most informative samples.
To address this challenge, we proceed to introduce a novel approach for facilitating \fram{} in practical Non-IID settings in~\cref{sec:method}.


\begin{figure}[t]
    \centering
    \includegraphics[width=1\linewidth]{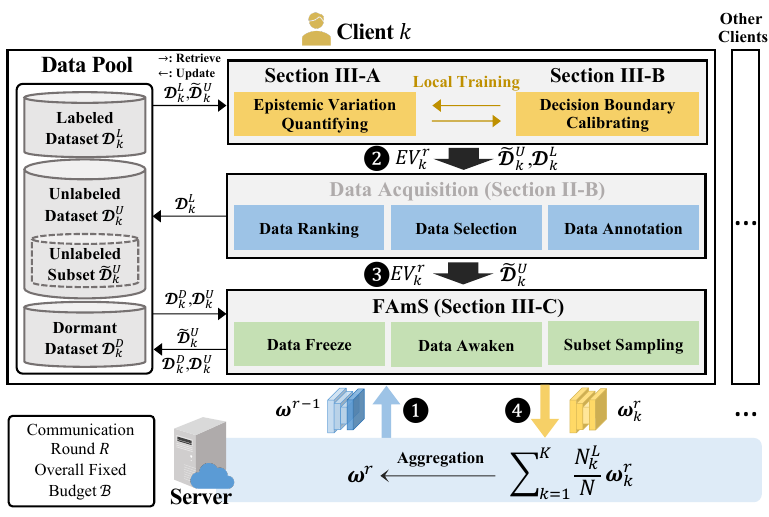}
    \caption{The workflow and integrated techniques of \model{}.}
    \label{fig:flow}
\end{figure}

\section{Proposed Techniques}
\label{sec:method}

To address the challenge of heterogeneity in \fram{}, we build upon the foundation laid in \cref{eq4}. Our approach involves two primary optimization strategies: 1) to optimize $\mathcal{D}_{k}^{S}$ by designing an acquisition strategy $A(\cdot)$ that considers client heterogeneity;
2) to optimize $\boldsymbol{\omega}_{k}^{r}$ by designing a new loss function $\ell(\cdot)$ to stabilize model parameter variation.

As depicted in \cref{fig:flow} for client-side training
, we propose a method for effectively quantifying the EV of each candidate sample.
The quantification is based on the fluctuation in the model's historical learning outcomes, as to be presented in \cref{subsec:CEV}.
Additionally, considering that the diversity of client data can introduce discrepancies into both local and global models, thereby affecting the accuracy of EV quantification, we enhance the loss function by adding an alignment term that is able to calibrate models' decision boundaries (to be detailed in~\cref{subsec:CDB}).
After the local training (\ding{183} in~\cref{fig:flow}), the acquisition strategy, optimized through quantified EVs, is employed to rank, select, and annotate the unlabeled samples (see~\cref{ssec:FAL-fram}).
Subsequent to data acquisition (\ding{184}), a data freeze and awaken mechanism with subset sampling (\FAMS{}) is executed to facilitate the efficiency of subsequent computations, as to be presented in~\cref{subsec:EO}.
Next, we will go through these techniques by presenting their implementations at the \fram{} round $r$ for clarity.


\subsection{Quantifying Epistemic Variations}
\label{subsec:CEV}
\begin{figure}[t]
    \centering
    \includegraphics[width=0.98\linewidth]{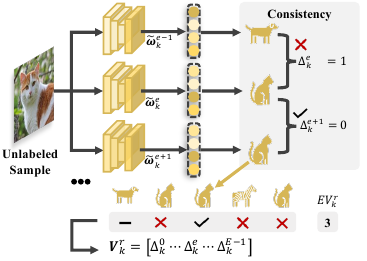}
    \caption{Example of quantifying the EV of an unlabled sample. For historical inference
    results [`dog', `cat', `cat', `zebra', `cat'], $\boldsymbol{V}=[0,1,0,1,1]$ and EV is calculated as $||\boldsymbol{V}||_0=3$.}
    \label{fig:Tloc}
\end{figure}

In a Non-IID setting, the frequent oscillations in the model's decision boundary are attributed to the varying distributions across clients. This suggests that relying solely on a single, potentially inaccurate state of the model for data selection, as seen in the current AL methods \cite{jia2019active, jin2022federated}, is often ineffective.

During the \fram{} training process, some samples are frequently swept by the fluctuating decision boundary, indicating their challenging nature for accurate model recognition. This characteristic observed in data samples is referred to as epistemic variation (EV).
Given this context, it is beneficial to involve human experts in addressing these challenging, high-EV unlabeled samples. Consequently, this motivates us to propose an acquisition strategy $A(\cdot)$ to effectively select unlabeled samples with high EVs.

Our strategy is based on the understanding that the fluctuating decision boundary of the model is mainly due to the aggregation of local models from heterogeneous clients into a single global model. This often results in a global model that differs significantly from the local models of previous rounds, leading to increased variability in local training.
Indeed, the ``cognition'' of an unlabeled sample is mirrored in its inferred label at the current model's decision boundary. 
Therefore, we can track the EVs of an unlabeled sample by observing the \emph{historical changes} in its inferred labels. Stable inferences imply consistent cognition for an unlabeled sample while fluctuating inferences indicate changing cognition.
Building on these insights, our strategy involves setting checkpoints during the local training phase. These checkpoints are designed to assess the consistency of sample inferences across adjacent training rounds, providing a means to quantify their EVs. 

Consider a local model $\boldsymbol{\omega}_{k}^{r}$ trained on the client $k$ for $E$ epochs. 
To quantify the EV for each candidate unlabeled sample, \model{} determines whether its inference results are \textbf{inconsistent} within each of two adjacent epochs.
A simple example is depicted in \cref{fig:Tloc}, where an unlabeled sample has been inferred for $E$ = 5 epochs by the local model. 
Three adjacent inference pairs (i.e., `dog' $\rightarrow$ `cat', `cat' $\rightarrow$ `zebra' and `zebra' $\rightarrow$ `cat' marked in red in~\cref{fig:Tloc}) show inconsistency, leading to an EV of 3. 
We proceed to present the details of EV computation, with the pseudocode listed in~\cref{alg:cev}.


\begin{algorithm}[!htbp]
\caption{\textsc{quantifyingEV} (local training epoch $E$, checkpoint of partially trained model $\boldsymbol{\tilde{\omega}}_{k}^{e}$)}
\label{alg:cev}
\begin{algorithmic}[1]
    \For{epoch $e \in (0,E]$}
        \For{each sample $\boldsymbol{x}_i \in \mathcal{D}_k^{U}$}
            \State $\hat{y}^{i,e}={\arg \max_{c}} p (y_{c}^{i} |\boldsymbol{\tilde{\omega}}_{k}^{e}  ;\boldsymbol{x}^{i})$ \Comment{inference on $\boldsymbol{\tilde{\omega}}_{k}^{e}$}
            \If{$e = 1$}
                $\Delta_{k}^{e} \gets 0$
            \Else
                ~$\Delta_{k}^{e}(\boldsymbol{\tilde{\omega}}_{k}^{e};\boldsymbol{x}^{i}) \gets \mathds{1}({\hat{y}^{i,e} \neq \hat{y}^{i,e-1}})$
            \EndIf
            \State $\boldsymbol{V}_{k}^{r}(\boldsymbol{x}^{i}).\text{append}(\Delta_{k}^{e})$
        \EndFor
    \EndFor
    \For{each sample $\boldsymbol{x}_i \in \mathcal{D}_k^{U}$} 
        \State ${EV}_{k}^{r}(\boldsymbol{x}^{i}) = ||\boldsymbol{V}_{k}^{r}(\boldsymbol{x}^{i})||_0$
    \EndFor
    \State  \Return $\{{EV}_{k}^{r}(\boldsymbol{x}^{i}) \mid \boldsymbol{x}^{i} \in \mathcal{D}_k^{U}\}$
\end{algorithmic}
\end{algorithm}

Let $\boldsymbol{\tilde{\omega}}_{k}^{e}$ be the checkpoint of the partially trained model after being updated for $e \in (0, E]$ epochs.
Thus, $\hat{y}^{i,e}={\arg \max_{c}} p (y_{c}^{i} |\boldsymbol{\tilde{\omega}}_{k}^{e} ;\boldsymbol{x}^{i})$
means the inference result for the sample $\boldsymbol{x}^{i}$ after $e$ epochs of training, where $p(y_{c}^{i} | \boldsymbol{\tilde{\omega}}_{k}^{e};\boldsymbol{x}^{i})\in [0,1]$ ($c \in \{1 \dots C\}$) denotes the output of $\boldsymbol{x}^{i}$ in the last layer's Softmax activation (see line~3 of~\cref{alg:cev}).

For each client's model $\boldsymbol{\tilde{\omega}}_{k}^{e}$, we make an inference on the unlabeled set $\mathcal{D}_{k}^{U}$ and record the results as $\hat{y}^{i,e}|_{i=1}^{N_{k}^{U}}$.
As in \cref{lo_variation}, the local model undergoes a variation $\Delta_{k}^{e}$ when two consecutive inferences of $\boldsymbol{x}^{i}$ are inconsistent (e.g., `dog' of $\boldsymbol{\tilde{\omega}}_{k}^{e-1}$ and `cat' of $\boldsymbol{\tilde{\omega}}_{k}^{e}$ in \cref{fig:Tloc}). 
We use an $E$-dimensional vector $\boldsymbol{V}$ to record the historical variations of inferences on each sample within $E$ epochs (see lines~5--6 of~\cref{alg:cev}).
The vector $\boldsymbol{V}_{k}^{r}(\boldsymbol{x}^{i})$ of client $k$ at round $r$ is computed as
\begin{align}
    &\boldsymbol{V}_{k}^{r}(\boldsymbol{x}^{i}) = [\Delta_{k}^{0}, \cdots, \Delta_{k}^{e}, \cdots, \Delta_{k}^{E-1}], \label{T_local}  \\ 
    &\Delta_{k}^{e}(\boldsymbol{\tilde{\omega}}_{k}^{e};\boldsymbol{x}^{i}) = \mathds{1}({\hat{y}^{i,e} \neq \hat{y}^{i,e-1}}), \label{lo_variation}
\end{align}
where $\mathds{1}$ is an indicator function, and the first element $\Delta_{k}^{0}=0$ (line~4).
Finally, the corresponding EV is measured as the number of 
inconsistent adjacent inference pairs, formally, ${EV}_{k}^{r}(\boldsymbol{x}^{i}) = ||\boldsymbol{V}_{k}^{r}(\boldsymbol{x}^{i})||_0$ (line~8)~\footnotemark. 
\footnotetext{We have explored more granular strategies of quantifying EVs, such as adding the class number and adding the EVs from the global model. However, our evaluation in \cref{sssec:exp_evquant} shows that simply using inference changes is sufficient in studied tasks.}

Therefore, we greedily select samples with the highest EVs in the new acquisition strategy $A(\cdot)$, instead of finding a batch of data to minimize the expected risk. In summary, for each client $k$, \cref{eq4} is rewritten as follows:
\begin{equation}
\label{eq7}
\begin{split}
&\min\nolimits_{|\mathcal{D}_{k}^{S}| \leq \mathcal{B}_{k}}\mathbb{E}[ 
\ell(\boldsymbol{\omega}^{r-1}; \mathcal{D}_{k}^{L} \cup \mathcal{D}_{k}^{S})], \\
& \mathcal{D}_{k}^{S} = \{\boldsymbol{x}^{i} \mid A({EV}_{k}^{r}(\boldsymbol{x}^{i}); \mathcal{B}_{k})\}.
\end{split}
\end{equation}
This approach is particularly advantageous for Non-IID scenarios where identifying the batch that minimizes global loss is challenging. 
Notably, our strategy considers the intrinsic complexity of samples within uniform client distributions as well as the epistemic variation arising from varied client distributions. This dual consideration guarantees that the strategy is applicable and efficient across both IID and Non-IID contexts, as studied in~\cref{ssec:exp_overall}.

\subsection{Calibrating Decision Boundaries}
\label{subsec:CDB}

For heterogeneous \fram{}, clients train the model as \cref{eq7} with two drawbacks.
One, each client focuses too much on its own data distribution and thus ignores global knowledge, resulting in samples that could be learned well after aggregation being quickly forgotten locally.
For another, the aggregation makes local training easily disturbed by other clients, thus diluting the local knowledge learned and resulting in samples learned in the previous round needing to be re-learned.
This dilemma of heterogeneous \fram{} aligns with the inefficiency problem in FL training under Non-IID, as emphasized by\text{~\cite{XuHHJ22,lee2022preservation,huang2022learnbe,xu2022lock}}, which stems from catastrophic forgetting during local training of knowledge encoded by other clients' models. While they address this by generating pseudo data locally~\text{\cite{XuHHJ22}} or knowledge distillation~\text{\cite{lee2022preservation,huang2022learnbe,xu2022lock}} to reinforce global knowledge, we tackle the issue from a data selection perspective, targeting samples that oscillate near multiple decision boundaries caused by Non-IID.
To prevent the \emph{selection of noisy samples} in resultant inaccurate models, we devise a new loss function to calibrate the models' decision boundaries along the training.

The general idea of decision boundary calibration is to introduce an additional loss term during local training, which aligns the representation of each sample with its counterpart at the previous training round.
However, in heterogeneous \fram{}, sample representations are learned both at the local and global models.
Hence, it is hard to determine whether a sample should be aligned using the local or the global model.

\begin{figure}[t]
    \centering
    \includegraphics[width=1\linewidth]{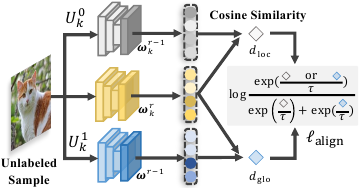}
    \caption{Example of the align loss term computation for an unlabeled sample.}
    \label{fig:align}
\end{figure}

In \model{}, each sample's EV is quantified as its 
difficulty by the corresponding local model (see \text{\cref{subsec:CEV}}).
Hence, we could utilize the EVs to instruct the model alignment:
For samples with low EV, the server's aggregation can amplify their EV in the next round, causing them to be mistakenly identified as high EV samples and subsequently selected, which negatively impacts the global model's performance. To address this, we align their representation in the current local model with that of the previously trained local model, as the latter has already learned to classify these samples accurately.
For samples with large EVs, their alignment is realized using the global model as the global model typically extracts better representations for its aggregation capability~\text{\cite{Qinbin2021Contra}}.
The necessity of these two alignment strategies for performance improvement is explored in detail in \text{\cref{sssec:exp_cdb}}.

The procedure of the align loss computation is depicted in~\cref{fig:align}.
Specifically, at round $r$, we first classify the unlabeled set into two categories $U_{k}^{j}$ ($j\in\{0,1\}$) according to whether ${EV}_{k}^{r-1}(\boldsymbol{x}^{i})$ at the ($r-1$)-th round is greater than the mean EV of all samples (denoted as $\overline{EV}$):
\begin{equation} \label{eq5}
\begin{aligned}
&U_{k}^{0} \leftarrow\left\{\boldsymbol{x}^{i} \mid {EV}_{k}^{r-1}\left(\boldsymbol{x}^{i}\right) \leq \overline{EV}\right\}, \\
&U_{k}^{1} \leftarrow\left\{\boldsymbol{x}^{i} \mid {EV}_{k}^{r-1}\left(\boldsymbol{x}^{i}\right) >\overline{EV}\right\}, \\
&\overline{EV}=\sum\nolimits_{i=1}^{N_{k}^{U}} {EV}_{k}^{r-1}\left(\boldsymbol{x}^{i}\right) / N_{k}^{U}.
\end{aligned}
\end{equation}

Subsequently, we denote the features of $\boldsymbol{x}^{i}$ (i.e., the output before Softmax layer) in the current model as $\boldsymbol{f}(\widetilde{\boldsymbol{\omega}}_{k}^{e}; \boldsymbol{x}^{i})$  ($\boldsymbol{x}^{i} \in U_{k}^{j}$).
Accordingly, we use the cosine similarity $cos(\cdot)$ to calculate $\boldsymbol{f}(\widetilde{\boldsymbol{\omega}}_{k}^{e}; \boldsymbol{x}^{i})$'s distance to the counterpart on the trained local model and on the trained global model, respectively.
We denote the two distances as $d_\text{loc}$ and $d_\text{glo}$ and compute them as in \cref{eq6}.
\begin{equation} \label{eq6}
\begin{aligned}
&d_\text{loc}(\boldsymbol{x}^{i}) = cos(\boldsymbol{f}(\widetilde{\boldsymbol{\omega}}_{k}^{e} ; \boldsymbol{x}^{i}), \boldsymbol{f}(\boldsymbol{\omega}_{k}^{r-1} ; \boldsymbol{x}^{i})), \\
&d_\text{glo}(\boldsymbol{x}^{i}) = cos(\boldsymbol{f}(\widetilde{\boldsymbol{\omega}}_{k}^{e} ; \boldsymbol{x}^{i}), \boldsymbol{f}(\boldsymbol{\omega}^{r-1} ; \boldsymbol{x}^{i})).
\end{aligned}
\end{equation}

Finally, we define the alignment loss term in \cref{eq10} in the Softmax form, aligning the representations of the samples with small (\emph{resp}. high) EVs using the local (\emph{resp}. global) model:
\begin{equation}\label{eq10}
\ell_{\text{align}}=-\mathbb{E}_{\boldsymbol{x}^{i} \in U_{k}^{j}}{[\log \frac{\exp(d_\text{*}(\boldsymbol{x}^{i})/\tau)}{\exp (d_\text{loc}(\boldsymbol{x}^{i})/\tau)+\exp(d_\text{glo}(\boldsymbol{x}^{i})/\tau)}}],
\end{equation}
where $\tau \in (0,1]$ denotes a temperature parameter following the study \cite{sohn2016improved}, and $d_\text{*}(\boldsymbol{x}^{i})$ means $d_\text{loc}(\boldsymbol{x}^{i})$ if $\boldsymbol{x}^{i} \in U_{k}^{0}$ and $d_\text{glo}(\boldsymbol{x}^{i})$ otherwise.
For each training epoch, we randomly sample data from the unlabeled set to calculate the alignment loss term, and the batch size is equal to the size of currently trained labeled data. Finally, the overall loss function is updated as
\begin{equation}
\ell = \ell_{\text{class}} + \mu \cdot \ell_{\text{align}},
\end{equation}
where $\ell_{\text{class}}$ is the original loss (i.e., $\ell(\boldsymbol{\omega}^{r-1}; \mathcal{D}_{k}^{L} \cup \mathcal{D}_{k}^{S})$ in \cref{eq7}) and $\mu$ is a tradeoff hyperparameter (evaluated in \cref{sssec:hyperparameters}). 


\subsection{Optimizing Efficiency of \model{}}
\label{subsec:EO}

\subsubsection{Complexity Analysis}
In our proposed approach, the \fram{} process involves model training for $(R\cdot E)$ epochs per labeled data sample.
Additionally, each client requires $E$ times the inference overhead to compute $\boldsymbol{V}_{k}^{r}$ and a decision boundary calibration 
process for each of its unlabeled samples per round.
To sum up, the worst-case overall complexity of our approach is 
$\mathcal{O}\left(R \cdot E \cdot \left(N^U + N^L\right)\right)$.
As labeled samples are significantly fewer than unlabeled ones in AL ($N^L \ll N^U$), we focus on reducing the number of unlabeled samples to enhance the efficiency of our approach.

\subsubsection{Data Freeze and Awaken Mechanism with Subset Sampling (\FAMS{})}

\begin{figure}[t]
    \centering
    \includegraphics[width=1\linewidth]{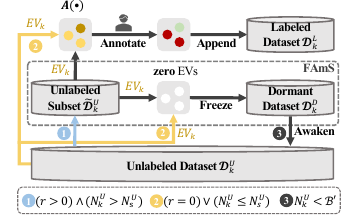}
    \caption{An illustration of \FAMS{}.}
    \label{fig:sample}
\end{figure}

In practice, model accuracy is usually dominated by only a small fraction of informative data samples.
For example, a study \cite{8057034} reports that a fraction of 50\% and 30\% of data samples can achieve 90\% model accuracy in the Wi-Fi localization and human activity recognition tasks, respectively.
This suggests that \fram{} should embrace possibly informative samples to make the whole process efficient.
To this end, we propose \FAMS{}, as depicted in \cref{fig:sample}.
The associated pseudocode is outlined in \cref{alg:FAMS}.
Specifically, before each local training, each client randomly\footnote{\FAMS{}'s insensitivity to `randomness' is verified in~\cref{ssec:exp_random}.} samples a \emph{subset} $\tilde{\mathcal{D}}_{k}^{U}$ of size $N_s^U$ from the unlabeled set $\mathcal{D}_{k}^{U}$ for the subsequent processes (see \ding{182} in \cref{fig:sample}).
However, for the initial round with $r = 0$ and when the unlabeled set is small enough such that $N_k^U \leq N_s^U$, the aforementioned subset sampling is waived (see \ding{183} in \cref{fig:sample}).
After the client $k$ trains locally based on the label set and the unlabeled subset, \model{} first performs data selection based on the acquisition strategy $A(\cdot)$. 
Then, for the remaining candidate unlabeled data,
\model{} freezes unlabeled samples with \textbf{zero} EVs (denoted as $\mathcal{D}_{k}^{U_0}$), and moves them to a dormant set $\mathcal{D}_{k}^{D}$ (lines~1--3 in~\cref{alg:FAMS}), i.e.,
\begin{align} \begin{aligned}
    &\mathcal{D}_{k}^{U_0} \leftarrow \{\boldsymbol{x}^{i} \in \tilde{\mathcal{D}}_{k}^{U} \setminus \mathcal{D}_{k}^{S} \mid {EV}_{k}^{r}(\boldsymbol{x}^{i}) = 0\}, \\
    &\mathcal{D}_{k}^{D} \leftarrow 
    \mathcal{D}_{k}^{D} \cup \mathcal{D}_{k}^{U_0}, \ \mathcal{D}_{k}^{U} \leftarrow
    \mathcal{D}_{k}^{U} \setminus (\mathcal{D}_{k}^{U_0} \cup \mathcal{D}_{k}^{S}).
    \label{zero-fluc}
\end{aligned} \end{align}

As \FAMS{} continues, the remaining unlabeled data samples diminish.
To ensure that there will always be enough unlabeled samples to be selected for annotation, we introduce the data awakening as follows (lines~4--5 in~\cref{alg:FAMS}).
Specifically, at the beginning of a new round, if the size of remaining unlabeled samples in $\mathcal{D}_{k}^{U}$ is smaller than a given threshold $\mathcal{B}'$ (see \ding{184} in \cref{fig:sample}), we awaken a random fraction $\beta$ of frozen data from $\mathcal{D}_{k}^{D}$ and move them to the unlabeled sample pool.
In the implementation, $\mathcal{B}'$ is set to be several times the annotation budget $\mathcal{B}_k$ per client.
The effect of the hyperparameter $\beta$ is evaluated in \cref{sssec:hyperparameters}.

Samples with zero EVs mainly come from two cases:
(1) the local model is overconfident and consistently misclassifies these samples;
(2) the local model consistently classifies the samples correctly as they are relatively simple.
Samples in both cases are relatively less useful to model training. Hence, it would be wise to temporarily not look at them to save the computational cost.
\FAMS{} significantly lowers the amount $N^U_k$ of samples that each client needs to infer in each round. 
Specifically, let $p_1$ and $p_2$ be the pruning ratios of subset sampling and data freeze, respectively. If discounting the minimal impact of data awaken, the optimal reduction in data volume can be expressed as $N^U_k \cdot p_1 \cdot p_2$.
Practically, $p_1$ is typically below 0.1 while $p_2$ hovers around 0.6, which can lead to a processing acceleration of an order of tens.

\begin{algorithm}[!htbp]
\caption{\FAMS{} (communication round $r$, epistemic variation matrix ${EV}_{k}^{r}$, sampled unlabeled subset $\tilde{\mathcal{D}}_{k}^{U}$, selected set $\mathcal{D}_{k}^{S}$, dormant set $\mathcal{D}_{k}^{D}$, unlabeled set $\mathcal{D}_{k}^{U}$)}
\label{alg:FAMS}
\begin{algorithmic}[1]
    \State $\mathcal{D}_{k}^{U_0} \leftarrow \{\boldsymbol{x}^{i} \in \tilde{\mathcal{D}}_{k}^{U} \setminus \mathcal{D}_{k}^{S} \mid {EV}_{k}^{r}(\boldsymbol{x}^{i}) = 0\}$ \Comment{Freeze}
    \State $\mathcal{D}_{k}^{D} \leftarrow 
    \mathcal{D}_{k}^{D} \cup \mathcal{D}_{k}^{U_0}$
    \State $\mathcal{D}_{k}^{U} \leftarrow \mathcal{D}_{k}^{U} \setminus (\mathcal{D}_{k}^{U_0} \cup \mathcal{D}_{k}^{S})$
    \If{$|\mathcal{D}_{k}^{U}| < \mathcal{B}'$} \Comment{Awaken}
        \State $\mathcal{D}_{k}^{U}.\text{append}(\texttt{RandomSample}(\mathcal{D}_{k}^{D},\beta|\mathcal{D}_{k}^{D}|))$
    \EndIf
    \If{$r > 0$ and $N_k^U \geq N_s^U$} \Comment{Subset Sampling}
        \State $\tilde{\mathcal{D}}_{k}^{U} \gets \texttt{RandomSample}(\mathcal{D}_{k}^{U},N_s^U)$
    \Else
        ~$\tilde{\mathcal{D}}_{k}^{U} \gets \mathcal{D}_{k}^{U}$
    \EndIf
\State \Return new unlabeled subset $\tilde{\mathcal{D}}_{k}^{U}$, updated unlabeled set $\mathcal{D}_{k}^{U}$, updated dormant set $\mathcal{D}_{k}^{D}$
\end{algorithmic}
\end{algorithm}

\section{Experiments}
\label{sec:experiment}
The code and instruction associated with our experiments are made available at GitHub~\cite{github-chase}.
\cref{ssec:settings} introduces the setups including the datasets, baselines, metrics, and parameter settings.
Comprehensive evaluations from \cref{ssec:exp_overall} to \cref{ssec:exp_sensitive} address the following key research questions:
\begin{itemize}[leftmargin=*]
\item \textbf{RQ1}: How does \model{} outperform the baselines across various data types and federation settings? (\cref{ssec:exp_overall})
\item \textbf{RQ2}:  
What is the impact of \model{}'s components on the model's effectiveness and efficiency, as determined by ablation studies? (\cref{ssec:exp_abla})
\item \textbf{RQ3:} How sensitive is our approach to variations in hyperparameters and the process randomness? (\cref{ssec:exp_sensitive}) 
\item \textbf{RQ4:} How does EV impact model accuracy, and is there a need for more nuanced EV metrics? (\cref{ssec:exp_evstudy})
\end{itemize}

\subsection{Experimental Settings} 
\label{ssec:settings}

\noindent
\textbf{Datasets.} We use five datasets for evaluations: {MNIST}, {EMNIST}, {CIFAR-10}, {CIFAR-100} and {Shakespeare}.

\noindent
\textbf{Baselines.} Our experiments involve comparisons with seven baselines: FedRandom, FedEntropy, FedCoreset, FedLL4AL, \fram{}($\bullet$), KAFAL, LoGo.

\noindent
\textbf{Metrics.} Effectiveness is measured by accuracy on the test set across a fixed number of communication rounds. Efficiency is assessed based on execution time and inference cost per \fram{} round.

\noindent
\textbf{FL-related Settings.}
clients $K=10$ for MNIST, CIFAR-10, and CIFAR-100, and $K=100$ for EMNIST and Shakespeare. Each client has $C_k=$ 5, 2, and 10 classes for CIFAR-10, MNIST, and EMNIST, respectively, simulating Non-IID settings. For CIFAR-100, data is partitioned using the Non-IID Dirichlet $Dir(\alpha=1e-2)$~\cite{Mikhail2019Bayesian}. We creating a
Non-IID dataset with 1,146 clients representing each speaking role in each play for Shakespeare~\cite{shakespeare} following~\cite{caldas2019leaf,mcmahan2017communication}.

\noindent
\textbf{AL-related Settings.} 
We initialize the labeled set by selecting small portions from the training set: 4\% for CIFAR-10 and Shakespeare, 10\% for CIFAR-100, 1.33\% for MNIST, and 2\% for EMNIST. Assuming honest, non-adversarial clients~\cite{ipeirotis2014repeated}, we examine two client behavior scenarios: 
For \textsf{AbCo} [\textsf{Absolute Cooperation}], clients are fully engaged with an equal annotation budget;
For \textsf{ReCo} [\textsf{Relative Cooperation}], annotation budgets vary based on client participation levels.


More comprehensive details on our setups can
be found in the supplementary material \text{\cite{github-chase}}.

\begin{figure*}[ht] 
\centering
\subfigure[MNIST / $C_k$=2 / \textsf{AbCo}]{ 
\begin{minipage}{0.25\linewidth}
\label{subfig:MNIST-GL-ACS-SAL}
    \includegraphics[
    width=4.2cm
    ]{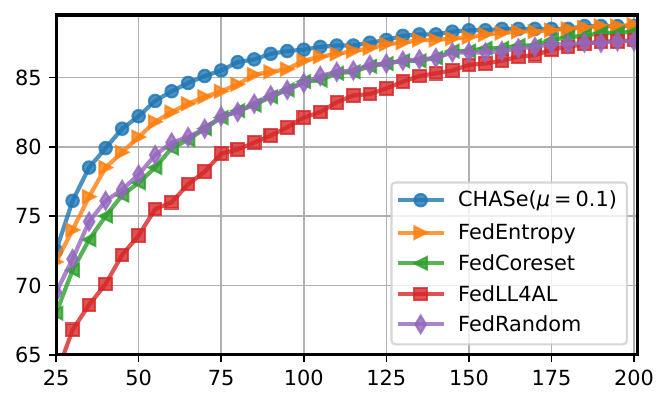}
\end{minipage}%
}%
\subfigure[MNIST / $C_k$=2 / \textsf{AbCo}]{
\begin{minipage}{0.25\linewidth}
    \label{subfig:MNIST-GL-ACS-FAL}
    \includegraphics[
    width=4.2cm
    ]{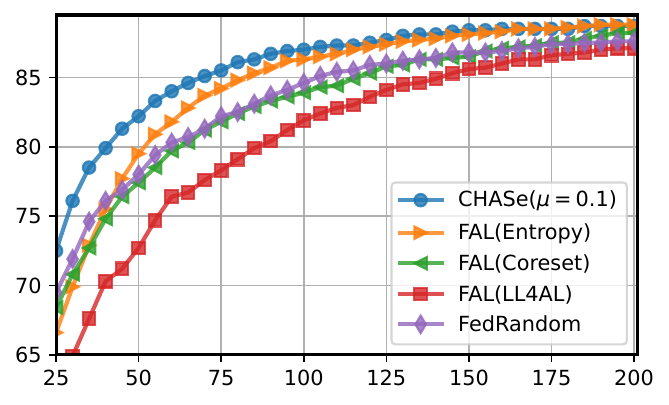}
\end{minipage}%
}%
\subfigure[CIFAR-10 / $C_k$=5 / \textsf{AbCo}]{ 
\begin{minipage}{0.25\linewidth}
\label{subfig:CIFAR10-GL-ACS-SAL}
    \includegraphics[
    width=4.2cm
    ]{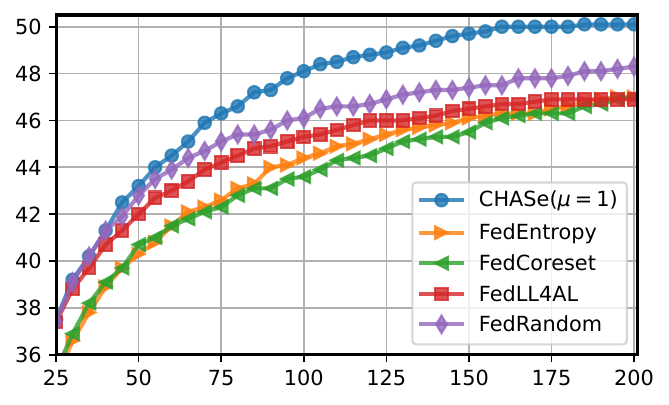}
\end{minipage}%
}%
\subfigure[CIFAR-10 / $C_k$=5 / \textsf{AbCo}]{
\begin{minipage}{0.25\linewidth}
    \label{subfig:CIFAR10-GL-ACS-FAL}
    \includegraphics[
    width=4.2cm
    ]{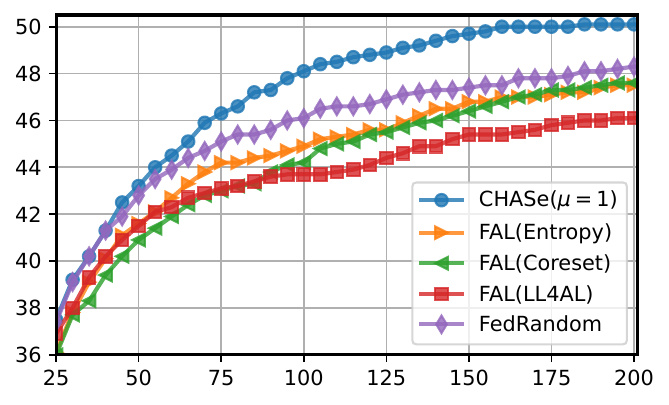}
\end{minipage}%
}%
\vspace{-0.0cm}
\subfigure[MNIST / $C_k$=2 / \textsf{ReCo}]{
\begin{minipage}{0.25\linewidth}
    \label{subfig:MNIST-GL-RCS-SAL}
    \includegraphics[
    width=4.2cm,
    ]{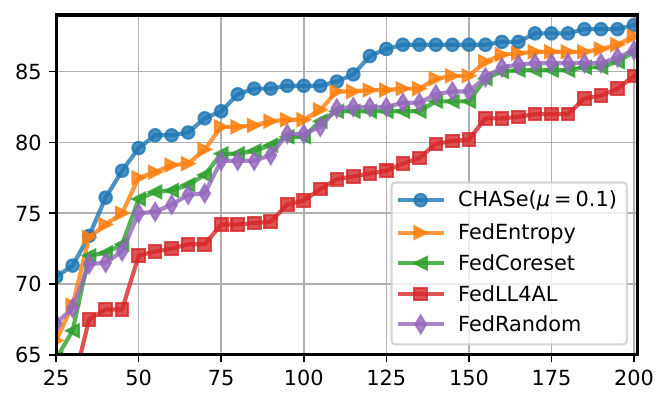}
\end{minipage}%
}%
\subfigure[MNIST / $C_k$=2 / \textsf{ReCo}]{
\begin{minipage}{0.25\linewidth}
    \label{subfig:MNIST-GL-RCS-FAL}
     \includegraphics[
    width=4.2cm
    ]{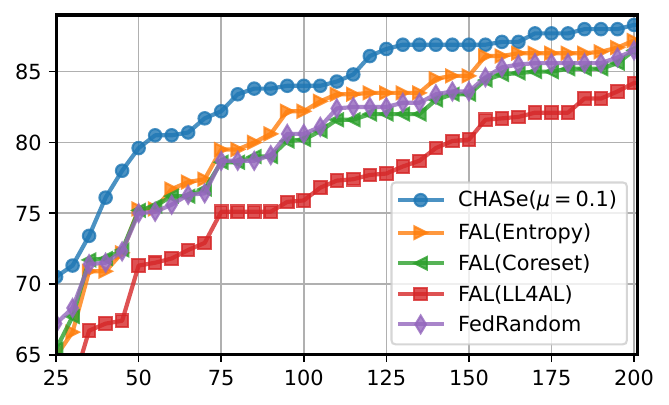}
\end{minipage}%
}%
\subfigure[CIFAR-10 / $C_k$=5 / \textsf{ReCo}]{
\begin{minipage}{0.25\linewidth}
    \label{subfig:CIFAR10-GL-RCS-SAL}
    \includegraphics[
    width=4.2cm
    ]{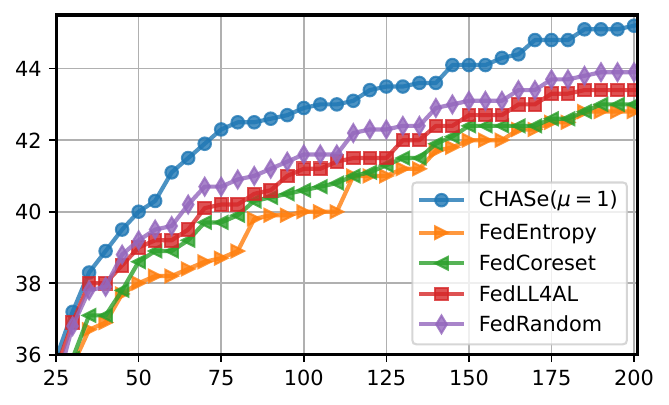}
\end{minipage}%
}%
\subfigure[CIFAR-10 / $C_k$=5 / \textsf{ReCo}]{
\begin{minipage}{0.25\linewidth}
    \label{subfig:CIFAR10-GL-RCS-FAL}
     \includegraphics[
    width=4.2cm
    ]{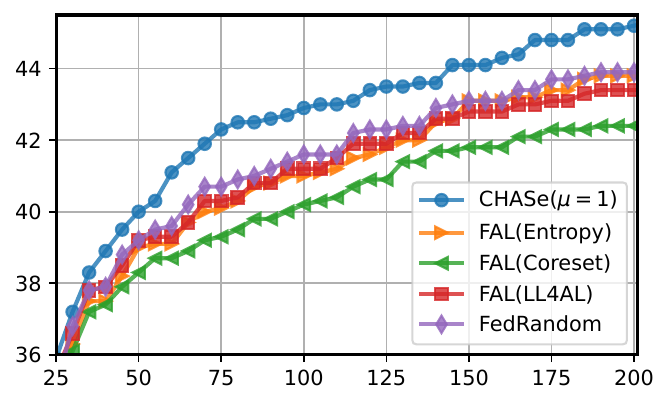}
\end{minipage}%
}%
\vspace{-0.0cm}
\subfigure[CIFAR-10 / IID / \textsf{ReCo}]{
\begin{minipage}{0.25\linewidth}
    \label{subfig:CIFAR10_RCS_IID_SAL}
    \includegraphics[width=4.2cm]{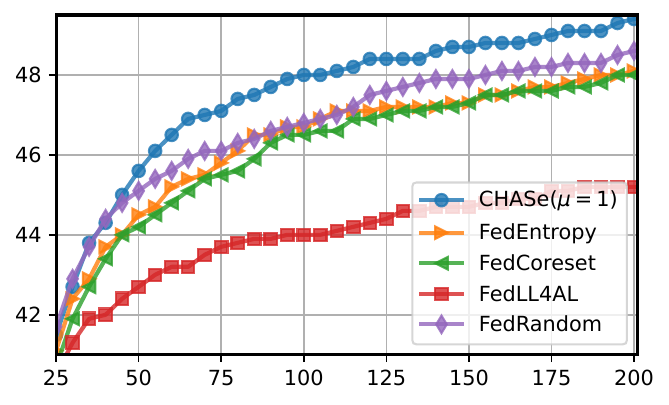}
\end{minipage}%
}%
\subfigure[CIFAR-10 / IID / \textsf{ReCo}]{
\begin{minipage}{0.25\linewidth}
    \label{subfig:CIFAR10_RCS_IID_FAL}
    \includegraphics[
    width=4.2cm
    ]{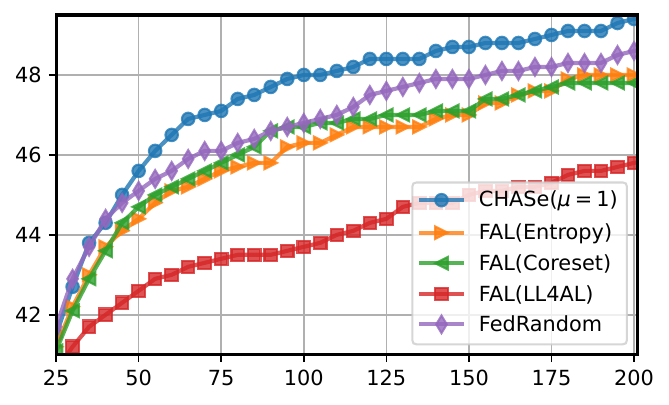}
\end{minipage}%
}%
\subfigure[CIFAR-100 / IID / \textsf{ReCo}]{
\begin{minipage}{0.25\linewidth}
    \label{subfig:CIFAR100_RCS_IID_SAL}
    \includegraphics[width=4.2cm]{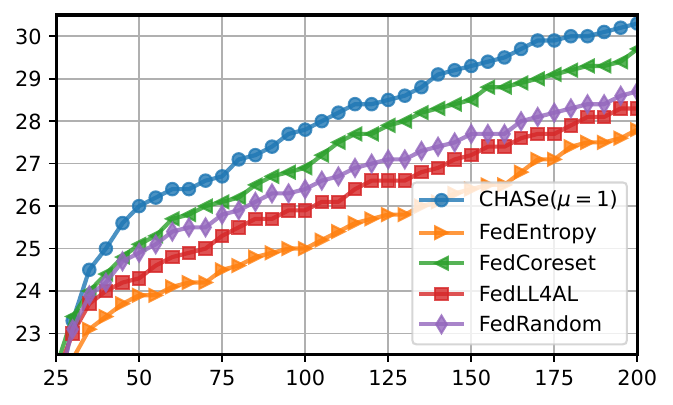}
\end{minipage}%
}%
\subfigure[CIFAR-100 / IID / \textsf{ReCo}]{
\begin{minipage}{0.25\linewidth}
    \label{subfig:CIFAR100_RCS_IID_FAL}
    \includegraphics[width=4.2cm]{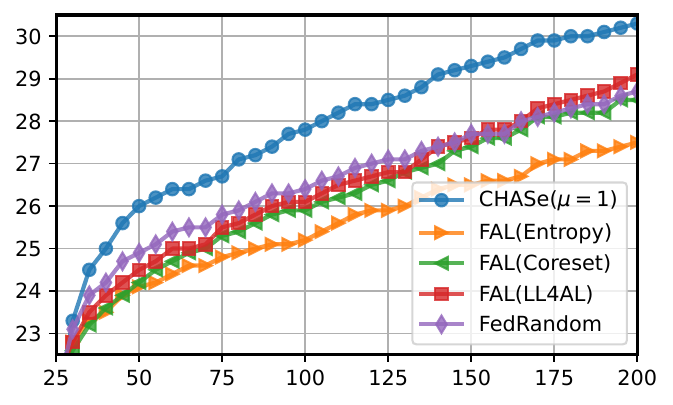}
\end{minipage}%
}%
\caption{Test accuracy (Y-axis) curves under \textsf{AbCo} and \textsf{ReCo} settings on different datasets. (a$\sim$d) are the accuracies of \model{} on \textsf{AbCo} and (e$\sim$h) are the counterparts on \textsf{ReCo}. (i$\sim$l) are the IID setting. The X-axis is the communication round in \fram{}.}
\label{fig:ACS-RCS-Non-IID}
\end{figure*}

\subsection{Overall Effectiveness Comparison (RQ1)}
\label{ssec:exp_overall}

Our overall comparison comprehensively considers the impact of different client behaviors (\textsf{AbCo} and \textsf{ReCo}), dataset types (image and text), and large data size and class number.

\subsubsection{\textsf{AbCo} vs Image Datasets}
We first look into the simpler image datasets MNIST and CIFAR-10.~\cref{subfig:CIFAR10-GL-ACS-SAL,subfig:CIFAR10-GL-ACS-FAL,subfig:MNIST-GL-ACS-SAL,subfig:MNIST-GL-ACS-FAL} present that \model{} always achieves the best accuracy scores in the \textsf{AbCo}.
%
When the communication round increases, the performance margins narrow down in MNIST, but \model{} still performs the best.

In \cref{subfig:MNIST-GL-ACS-SAL}, the advantage of \model{} (blue line) on the simpler dataset MNIST is more evident when fewer rounds (e.g., 75) are used.
When using the harder dataset CIFAR-10 (see \cref{subfig:CIFAR10-GL-ACS-SAL}), \model{} steadily outperforms FedEntropy and FedCoreset by around 2\% from the 50-th rounds onwards.
\cref{subfig:CIFAR10-GL-ACS-FAL,subfig:MNIST-GL-ACS-FAL} compare \textbf{the baselines in the \fram{}($\bullet$) family}. 
The aggregation of heterogeneous clients leads to the inaccurate decision boundary of the global model, so the boost of the \fram{} scheme \cite{jin2022federated} is limited.
Even \fram{}(LL4AL), the best among the \fram{}($\bullet$) family, performs poorly on CIFAR-10. In contrast, \model{} achieves stable performance due to its unique effectiveness techniques of quantifying EVs and calibrating decision boundaries.

\subsubsection{\textsf{ReCo} vs Image Datasets}
We further conduct experiments in the challenging \textsf{ReCo} setting.
Clients in \textsf{ReCo} have \textit{diverse annotation behaviors}, resulting in more heterogeneous data distributions among clients, which are reflected in the more erratic overall performance.
Our subsequent experiments will mainly focus on this more complex \textsf{ReCo} scenario.

As shown in \cref{subfig:CIFAR10-GL-RCS-SAL,subfig:CIFAR10-GL-RCS-FAL,subfig:MNIST-GL-RCS-SAL,subfig:MNIST-GL-RCS-FAL}, traditional AL methods (i.e., entropy-based, Core-set and LL4AL), which select samples only based on the current model state, do not perform well and sometimes even worse than random selection. 
We find that FedLL4AL is more susceptible to heterogeneous clients 
since its employed loss prediction also suffers from epistemic variation.
In contrast, \model{} achieves clear improvements and outperforms other baselines by over 2\% at 150 rounds.
These results verify the effectiveness of \model{} in combating the samples' 
epistemic variation in Non-IID.

\subsubsection{Image Datasets of Large Data Size \& Class Number}
We further test the accuracy of the methods on the two larger datasets with \textit{high numbers of classes}: EMNIST, which has 62 classes and a large number of data samples from 100 \textit{clients at a larger scale}, and CIFAR-100, which has 100 classes.

\begin{figure} 
\centering
\subfigure[CIFAR-100 / Dir(1e-2) / \textsf{ReCo}]{ 
\begin{minipage}{0.5\linewidth}
\label{subfig:CIFAR100-GL-RCS-SAL}
    \includegraphics[
    width=4.2cm,
    ]{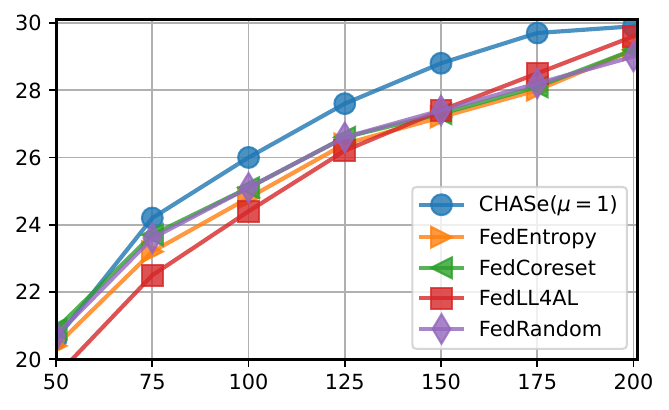}
\end{minipage}%
}%
\subfigure[EMNIST / $C_k$=10 / \textsf{ReCo}]{
\begin{minipage}{0.5\linewidth}
    \label{subfig:EMNIST-GL-RCS-SAL}
    \includegraphics[
    width=4.2cm,
    ]{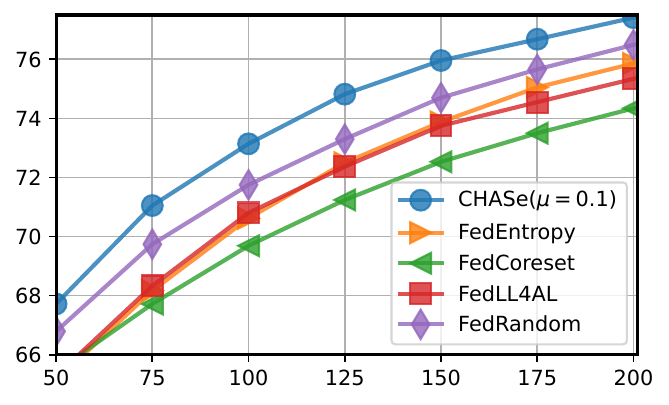}
\end{minipage}%
}%
    \caption{Test accuracy curves on CIFAR-100 and EMNIST.}
\end{figure}

As depicted in \cref{subfig:CIFAR100-GL-RCS-SAL}, \model{} shows a significant advantage in the middle to late stages, even on the more challenging CIFAR-100 dataset with more classes.
FedLL4AL slightly improves performance over \cref{subfig:CIFAR10-GL-RCS-SAL,subfig:CIFAR10-GL-RCS-FAL} on simpler CIFAR-10 because its employed loss prediction can now better extract features under the more complex VGG-19 model.
The fact that our approach dominates under multiple models (CNN\cite{xie2019dba} and VGG-19) indicates that it is to some extent model-agnostic.
Furthermore, we increase the number of clients with the local models fixed in the larger EMNIST dataset. As depicted in \cref{subfig:EMNIST-GL-RCS-SAL}, \model{} continues to offer a significant performance boost.
\model{} also clearly outperforms the \fram{}($\bullet$) family in the same settings. These results are omitted for the presentation brevity.
In summary, \model{} remains effective as the number of sample classes and clients increases.

\subsubsection{IID vs Image Datasets} \label{sssec:exp_iid}

We also validate the efficacy of \model{} on CIFAR-10 and CIFAR-100 datasets under the IID setting (i.e.,~clients have the same classes and equal numbers of samples per class), with the client's annotation behavior set to \textsf{ReCo}. The accuracy results are depicted in \cref{subfig:CIFAR10_RCS_IID_SAL,subfig:CIFAR10_RCS_IID_FAL,subfig:CIFAR100_RCS_IID_SAL,subfig:CIFAR100_RCS_IID_FAL}, which shows that our proposed method outperforms the other methods on both datasets. This demonstrates that \model{} not only achieves remarkable scores under Non-IID scenarios but also maintains its validity under IID.

Moreover, we observe that FedLL4AL exhibits poor performance on CIFAR-10, as shown in \cref{subfig:CIFAR10_RCS_IID_SAL,subfig:CIFAR10_RCS_IID_FAL}. The reason behind this is the shallow architecture of the network we used, which makes it difficult for the loss module of FedLL4AL to extract useful features. In \textsf{ReCo}, the client's annotation behavior causes more severe changes in the data distribution, leading to a more inaccurate global model used within \fram{}. This results in unstable and limited performance gains while using \fram{} in different methods.

We present an in-depth discussion on the effectiveness of our proposed \model{} in both IID and Non-IID scenarios for \fram{}. A key aspect of effective AL is the selection of informative samples, particularly those located near decision boundaries, which are more likely to be misclassified. However, selecting samples from a single model's decision boundary may not always be optimal due to model inaccuracies that are exacerbated in Non-IID scenarios.
To address this limitation, we extend the traditional AL approach to multiple decision boundaries by utilizing inference changes (EVs) from historical records to select informative samples that oscillate on these boundaries. Furthermore, we incorporate the alignment loss term to reduce local and global model inaccuracies, improving \fram{} performance.
Our proposed method effectively captures the complexities of multiple decision boundaries, making it a promising approach to selecting informative samples and improving \fram{} performance in both IID and Non-IID scenarios.

\begin{figure}[!t] 
\centering
\subfigure[Client Data Distribution]{ 
    \begin{minipage}{0.5\linewidth}
    \label{fig:text_dis}
        \includegraphics[
        width=4.2cm,
        ]{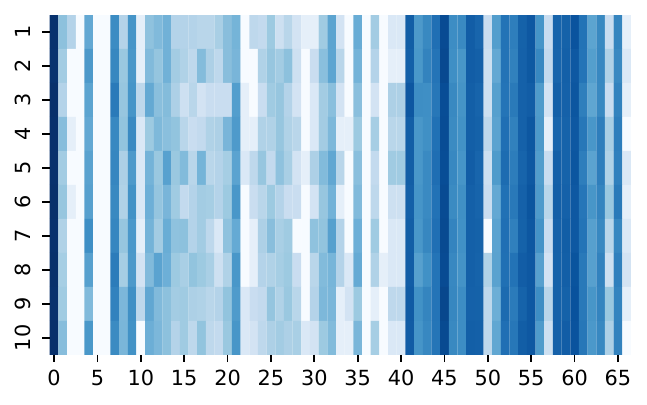}
    \end{minipage}%
    }%
    \subfigure[Shakespeare / By Role / \textsf{ReCo}]{
    \begin{minipage}{0.5\linewidth}
        \label{fig:text_exp}
        \includegraphics[
        width=4.2cm,
        ]{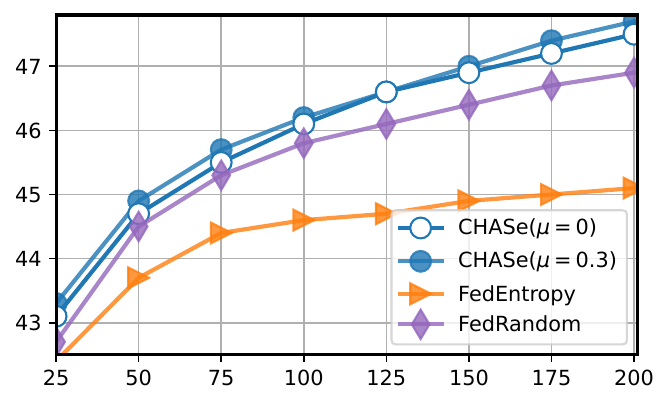}
    \end{minipage}%
    }%
    \caption{Data distribution and test accuracy on Shakespeare.}
\end{figure}
\subsubsection{\textsf{ReCo} vs Text Dataset} \label{sssec:textexp}

We continue to explore the effectiveness of \model{} in the text domain.
\cref{fig:text_exp} plots the distribution of samples among 10 selected from the 100 clients (represented on Y-axis) over 80 classes (X-axis).
This figure uses a logarithmic scale to represent data quantity, with darker colors indicating more data per class.
There is a notable variety in the data distribution among different clients in this federation.
Based on the evaluation in \cref{subfig:CIFAR10-GL-RCS-SAL,subfig:CIFAR10-GL-RCS-FAL,subfig:MNIST-GL-RCS-SAL,subfig:MNIST-GL-RCS-FAL} under Non-IID and \textsf{RCS}, we include FedRandom and FedEntropy for comparisons as they outstand as the second-best method on CIFAR-10 and MNIST, respectively.
\cref{fig:text_exp} shows that \model{} achieves the optimal convergence efficiency in the next character prediction under \textsf{ReCo}, highlighting its generalizability in the other domain.
Additionally, we find that the \model{} instance with a decision boundary calibration ($\mu=0.3$) outperforms the one without this technique ($\mu=0$). 
This finding preliminarily confirms the effectiveness of each component of our performance.
Since image tasks are more challenging and align with prior research \cite{9835327,9835537}, our follow-up experiments mainly focus on image datasets.

\subsection{Ablation Study (RQ2)}
\label{ssec:exp_abla}

\subsubsection{Effectiveness Techniques}

We design experiments in \cref{tab:ablation} to investigate the effectiveness of the techniques of quantifying EVs (denoted as EV) and calibrating decision boundaries (denoted as $\ell_{\text{align}}$).
If none of these two techniques are used, \model{} degenerates to FedRandom.

\cref{tab:ablation} shows that quantifying EVs is effective in improving accuracy across datasets.
Measured as the inconsistency of adjacent inference results, the EV indicates the difficulty level of the sample being inferred.
Therefore, utilizing EVs to select unlabeled samples helps improve models' accuracy in \fram{}.
Furthermore, \cref{tab:ablation} shows that adding the alignment loss $\ell_{\text{align}}$ to the model training further improves accuracy, particularly on challenging datasets like CIFAR-100. 
This finding indicates that considering only local distributions during client model updating is insufficient, and employing the newly designed loss term leads to a more accurate model, which prevents the selection of noisy samples and avoids unnecessary expenses. 
Note that EMNIST has abundant data and more clients and thus the alignment loss term may not help much with the accuracy.
The supplementary material \text{\cite{github-chase}} provides clearer visualizations showing that \model{}, with or without alignment loss, outperforms all baselines.

\begin{table}[t]
\caption{Ablation study of EV and $\ell_{\text{align}}$ under \textsf{ReCo}.}
\footnotesize
\renewcommand{\arraystretch}{0.5}
\setlength{\tabcolsep}{1.5mm}{
\begin{tabular}{@{}c|cc|ccccccc@{}}
\toprule
\multicolumn{3}{c|}{\model{}} & \multicolumn{7}{c}{Test Accuracy (\%) at Round $r$}                                  \\ \midrule
Datasets & EV &$\ell_{\text{align}}$ &50 &75 &100 &125 &150 &175 &200           
\\ 
\midrule
\multirow{3}{*}{MNIST}     &            &           &75.0 &78.7 &80.6 &82.5 &83.6 &85.6 &86.5 \\
                           &+  &           &78.9 &82.2 &83.8 &85.0 &86.5 &87.6 &87.8 \\
                           &+  &+ &\textbf{79.6} &\textbf{82.2} &\textbf{84.0} &\textbf{86.6} &\textbf{86.9} &\textbf{87.7} &\textbf{88.3} \\ 
\midrule
\multirow{3}{*}{EMNIST}   &            &          &66.8 &69.7 &71.8 &73.3 &74.7 &75.7 &76.5 \\
                           &+  &         &67.7 &71.0 &\textbf{73.3} &74.7 &76.0 &76.7 & 77.4  \\
                           &+  &+ 
                           &\textbf{67.7} &\textbf{71.1} &73.1 &\textbf{74.8} &\textbf{76.0} &\textbf{76.7} &\textbf{77.7} \\ 
                           
\midrule
\multirow{3}{*}{CIFAR-10}  &           &           &39.2 &40.7 &41.6 &42.3 &43.1 &43.7 &43.9  \\
                           &+ &           &39.8 &41.3 &42.6 &\textbf{43.6} &43.7 &44.8 &44.8 \\
                           &+ &+ &\textbf{40.0} &\textbf{42.3} &\textbf{42.9} &43.5 &\textbf{44.1} &\textbf{44.8} &\textbf{45.2} \\ 
\midrule
\multirow{3}{*}{CIFAR-100} &            &           
                        &20.7 &23.6 &25.1 &26.6 &27.4 &28.2 &29.0  \\
                           &+  &           &\textbf{21.1} &23.8 &25.6 &26.9 &28.1 &29.0 &\textbf{30.0}  \\
                           &+  &+   &20.6 &\textbf{24.2} &\textbf{26.0} &\textbf{27.6} &\textbf{28.8} &\textbf{29.7} &29.9 \\ 
\bottomrule
\end{tabular}}
\label{tab:ablation}
\end{table}

\begin{figure*}[th]
\centering
     \subfigure[MNIST / $C_k$=2 / \textsf{ReCo}]{
    \begin{minipage}{0.24\linewidth}
        \label{subfig:NPR-MNIST}
        \includegraphics[width=4.2cm]{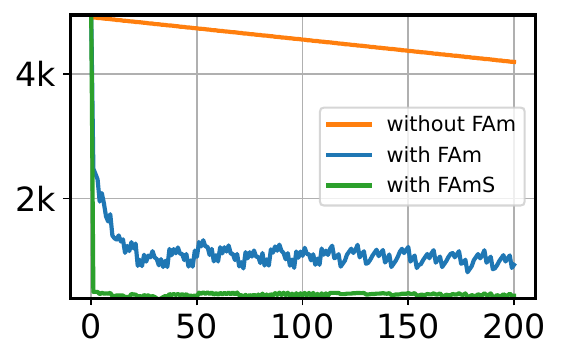}
    \end{minipage}
    }%
   \subfigure[EMNIST / $C_k$=10 / \textsf{ReCo}]{
    \begin{minipage}{0.24\linewidth}
        \label{subfig:NPR-EMNIST}
        \includegraphics[width=4.2cm]{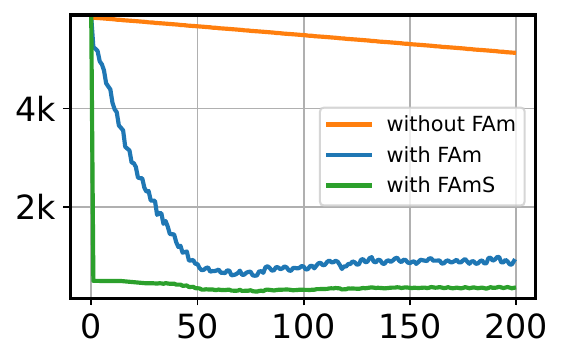}
    \end{minipage}
    }%
     \subfigure[CIFAR-10 / $C_k$=5  / \textsf{ReCo}]{
    \begin{minipage}{0.24\linewidth}
        \label{subfig:NPR-CIFAR10}
        \includegraphics[width=4.2cm]{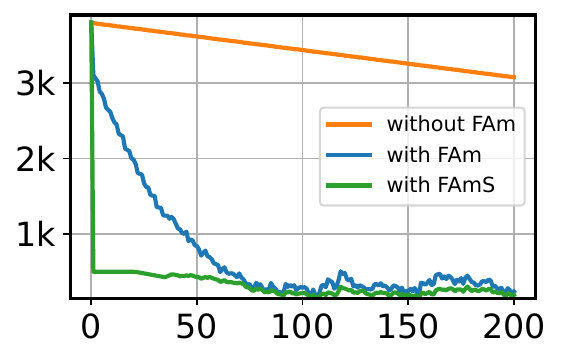}
    \end{minipage}
    }%
    \subfigure[CIFAR-100 / Dir(1e-2)  / \textsf{ReCo}]{
    \begin{minipage}{0.24\linewidth}
    \label{subfig:NPR-CIFAR100}
        \includegraphics[width=4.2cm]{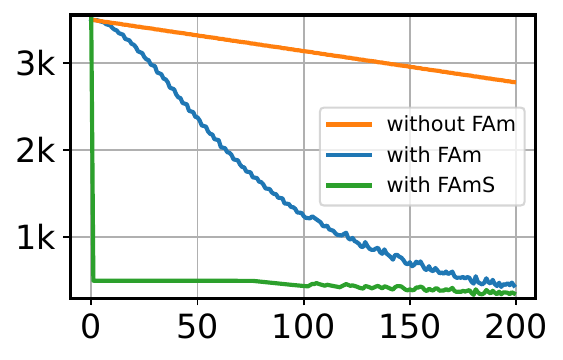}
    \end{minipage}
    }%
    \vspace{-0.3cm}
    \caption{The average inference cost per client for \model{}'s variants.}
    \label{fig:SV}
\end{figure*}

\subsubsection{Efficiency Technique}
\label{sssec:exp_efficiency}

Here, we validate the impact of using \FAMS{}.
In particular, we measure the efficiency in terms of the size of inferred unlabeled samples for three \model{} variants: one without the freeze and awaken mechanism (\textsf{FAm}), one with \textsf{FAm}, and one with \FAMS{}.
The results, as shown in \cref{fig:SV}, show a clear decrease in average inference cost when applying the data freeze and awaken mechanism (with \textsf{FAm} and with \FAMS{}).
Additionally, in 
challenging datasets such as CIFAR-100 (as seen in \cref{subfig:NPR-CIFAR100}), subset sampling in \FAMS{} significantly enhances efficiency gains.
%

Finally, we scrutinize the tradeoff between accuracy and time cost when employing the alignment loss $\ell_{\text{align}}$.
\cref{tab:P&C} shows that the full version of \model{} ($\mu=1$) with $\ell_{\text{align}}$ achieves the highest accuracy, although at the expense of some efficiency.
Without using $\ell_{\text{align}}$, \model{} ($\mu=0$) saves time but results in lower accuracy.
Nonetheless, \model{} ($\mu=0$) outperforms most of the baselines on accuracy and time cost.
FedCoreset (without inference to the classification layer) and FedRandom (without any inference) run faster than \model{} ($\mu=0$), but their accuracy is much lower.

\begin{table}
\centering
\caption{Accuracy vs time cost on CIFAR-10 under \textsf{ReCo}.}
\footnotesize
\renewcommand{\arraystretch}{0.6}
\setlength{\tabcolsep}{1.2mm}{
\begin{tabular}{c|*{7}{S}|c@{}}
\toprule
\multirow{2}{*}{Methods} & \multicolumn{7}{c|}{Test Accuracy (\%) at Round $r$} & \multirow{2}{*}{\begin{tabular}[c]{@{}c@{}}Time Cost \\ (min/r)\end{tabular}} \\ \cmidrule(r){2-8}
& {50} & {75} & {100} & {125} & {150} & {175} & {200} & \\
\midrule
\model{} ($\mu=1$) &\textbf{40.0} &\textbf{42.3} &\textbf{42.9} &\uline{43.5} &\textbf{44.1} &\textbf{44.8} &\textbf{45.2} & 2.0 \\
\model{} ($\mu=0$) & \uline{39.8} &\uline{41.3} &\uline{42.6} &\textbf{43.6} &\uline{43.7} &\textbf{44.8} &\uline{44.8} & 0.99 \\
FedEntropy & 38.0 & 38.7 & 40.0 & 41.0 & 42.0 & 42.5 & 42.8 & 1.0 \\
\fram{}(Entropy) & 39.0 & 40.1 & 41.0 & 41.8 & 43.1 & 43.4 & 43.8 & 1.16 \\
FedCoreset & 38.6 & 39.7 & 40.6 & 41.3 & 42.4 & 42.6 & 43.0 & 0.66 \\
\fram{}(Coreset) & 38.3 & 39.3 & 40.2 & 40.9 & 41.8 & 42.3 & 42.4 & 0.67 \\
FedLL4AL & 39.0 & 40.2 & 41.2 & 41.5 & 42.7 & 43.3 & 43.4 & 1.08 \\
\fram{}(LL4AL) & 39.2 & 40.3 & 41.2 & 41.9 & 42.8 & 43.1 & 43.4 & 0.99 \\
FedRandom & 39.2 & 40.7 & 41.6 & 42.3 & 43.1 & 43.7 & 43.9 & \textbf{0.63} \\
\bottomrule
\end{tabular}}
\label{tab:P&C}
\end{table}

\begin{table}
\caption{Effect of federated learning on \textsf{ReCo}.}
\centering
\footnotesize
\renewcommand{\arraystretch}{0.6}
\setlength{\tabcolsep}{1.2mm}{
\begin{tabular}{@{}cc|*{7}{S}@{}}
\toprule
\multicolumn{2}{c|}{\multirow{2}{*}{Settings}}  & \multicolumn{7}{c}{Test Accuracy (\%) at Round $r$} \\ 
\cmidrule(r){3-9}
\multicolumn{2}{c|}{}
& \multicolumn{1}{c}{50}    & \multicolumn{1}{c}{75}    & \multicolumn{1}{c}{100}   & \multicolumn{1}{c}{125}   & \multicolumn{1}{c}{150}   & \multicolumn{1}{c}{175}  & \multicolumn{1}{c}{200}        \\
\midrule
\multicolumn{1}{c|}{\multirow{2}{*}{MNIST}}    & \textsf{isolated}
&31.5 &32.3 &33.0 &34.0 &35.0 &35.0 &35.0 \\
\multicolumn{1}{c|}{}    & \textsf{federation} [ours] 
&\textbf{79.6} &\textbf{82.2} & \textbf{84.0} &\textbf{86.6} &\textbf{86.9} &\textbf{87.7} &\textbf{88.3} \\ \hline
\multicolumn{1}{c|}{\multirow{2}{*}{CIFAR-10}} & \textsf{isolated} 
& 26.6 & 28.3 & 29.4 & 29.4 & 30.0 & 30.5 & 31.0 \\
\multicolumn{1}{c|}{}    & \textsf{federation} [ours]
&\textbf{40.0} &\textbf{42.3}  &\textbf{42.9}  &\textbf{43.5} &\textbf{44.1}  &\textbf{44.8}  &\textbf{45.2} \\
\bottomrule
\end{tabular}
}
\label{tab:noFL}
\end{table}

\subsubsection{Isolated AL vs Federated AL}
\label{sssec:necessity_fal}

Given Non-IID data distributed across clients, we design an \textsf{isolated} approach in which each client independently executes \model{} without the server's aggregation.
We compared the average accuracy of all clients using the \textsf{isolated} approach with the original \textsf{federation} approach in \cref{tab:noFL}. Our observations indicate that a single client faces difficulties in obtaining a high-quality model due to limited data and annotation budgets. 
Especially under the simple MNIST, the model trained by a single client fails to achieve good generalization performance.
These results highlight the need for a federated AL paradigm.

\begin{figure*}[ht]
\centering
     \subfigure[MNIST / $C_k$=2 / $\mu$]{
    \begin{minipage}{0.24\linewidth}
        \label{subfig:mu-MNIST}
        \includegraphics[width=4.2cm]{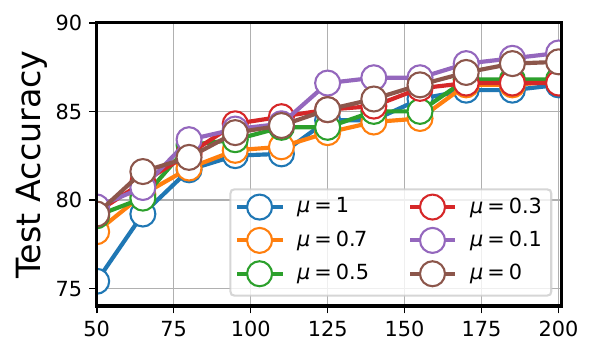}
    \end{minipage}
    }%
   \subfigure[CIFAR-10 / $C_k$=5 / $\mu$]{
    \begin{minipage}{0.24\linewidth}
        \label{subfig:mu-CIFAR10}
        \includegraphics[width=4.2cm]{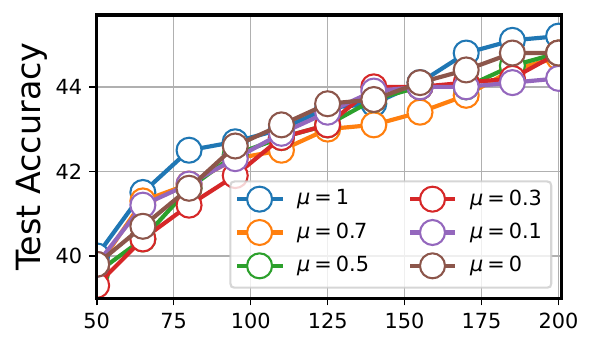}
    \end{minipage}
    }%
    \subfigure[CIFAR-10 / $C_k$=5 / $\beta$]{
    \begin{minipage}{0.24\linewidth}
    \label{subfig:Ta-CIFAR10-NPR}
        \includegraphics[width=4.2cm]{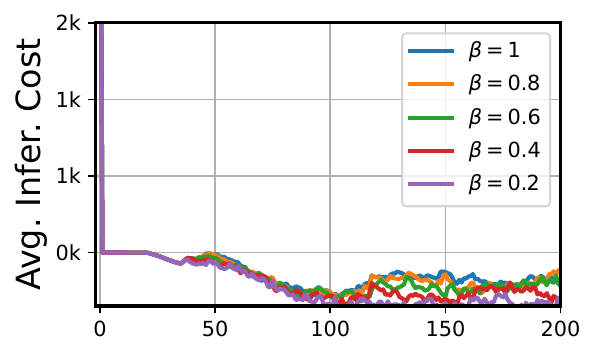}
    \end{minipage}
    }%
    \subfigure[CIFAR-10 / $C_k$=5 / $\beta$]{
    \begin{minipage}{0.24\linewidth}
        \label{subfig:Ta-CIFAR10-Perform}
        \includegraphics[width=4.2cm]{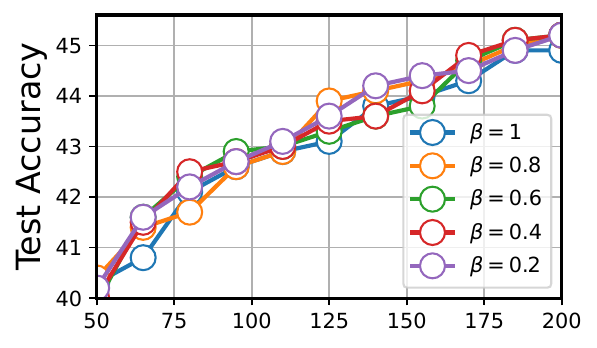}
    \end{minipage}
    }%
    \vspace{-0.3cm}
    \caption{Varying tradeoff parameter $\mu$ (see a$\sim$b) and awaken ratio $\beta$ (see c$\sim$d) under \textsf{ReCo}.}
    \label{fig:exp_hyper}
\end{figure*}

\subsection{Sensitivity Analysis (RQ3)}
\label{ssec:exp_sensitive}

\subsubsection{Effect of Hyperparameters}
\label{sssec:hyperparameters}

Our study focuses on two hyperparameters: the tradeoff parameter $\mu$ for the alignment loss term and the awaken ratio $\beta$ in \FAMS{}. 

\noindent\textbf{Tradeoff Parameter $\mu$}. Varying $\mu$ from 0 to 1, \cref{subfig:mu-MNIST,subfig:mu-CIFAR10} report the accuracy of MNIST and CIFAR-10, respectively.
An appropriate weight $\mu$ for the alignment loss term \emph{does} contribute to the accuracy of \model{}. 
On the simple MNIST dataset, as shown in \cref{subfig:mu-MNIST}, since the model learns very quickly, that is, the decision boundary is relatively stable, and the optimized acquisition strategy can already select high-informative samples more accurately.
Hence, a relatively small $\mu$, such as $\mu=0.1$, suffices. However, for a challenging dataset like CIFAR-100, a larger $\mu=1$ performs better as it can promptly reduce the changes of the decision boundary, as shown in  \cref{subfig:mu-CIFAR10}.

\noindent\textbf{Awaken Ratio $\beta$}. Varying $\beta$ from 0.2 to 1, \cref{subfig:Ta-CIFAR10-NPR,subfig:Ta-CIFAR10-Perform} report the  
average inference cost and the corresponding accuracy on CIFAR-10, respectively.
It is observed that increasing $\beta$ gradually raises the average inference cost while having little impact on the accuracy.
Moreover, since the probability of all clients awakening the dormant set simultaneously is low, there is still a significant cost improvement when using $\beta=1$ compared to without using \FAMS{} at all (see \cref{fig:SV}).

Generally, \model{} is \emph{insensitive} to both hyperparameters.

\begin{table}[t]
\caption{Effect of randomness in \FAMS{} on CIFAR-10 and CIFAR-100 under \textsf{ReCo} / Non-IID.}
\centering
\footnotesize
\renewcommand{\arraystretch}{1}
\setlength{\tabcolsep}{1.4mm}{
\begin{tabular}{@{}cc|*{7}{S}@{}}
\toprule
\multicolumn{2}{c|}{\multirow{2}{*}{Settings}}  & \multicolumn{7}{c}{Test Accuracy (\%) at Round $r$} \\ 
\cmidrule(r){3-9}
\multicolumn{2}{c|}{}
& \multicolumn{1}{c}{50}    & \multicolumn{1}{c}{75}    & \multicolumn{1}{c}{100}   & \multicolumn{1}{c}{125}   & \multicolumn{1}{c}{150}   & \multicolumn{1}{c}{175}  & \multicolumn{1}{c}{200}        \\
\midrule
\multicolumn{1}{c|}{}    & w/o \textsf{S} 
&\uline{78.4} &\textbf{83.9}  &\uline{85.3} &86.1 &86.8 &86.9 &87.7 \\
\multicolumn{1}{c|}{MNIST}    & w/o \textsf{FAm} 
&78.3 &\uline{82.2}  &\textbf{85.4}  &\textbf{86.7} &\textbf{87.1}  &\uline{87.5}  &\uline{88.2} \\
\multicolumn{1}{c|}{}    & w \FAMS{} [ours] 
&\textbf{79.6} &\uline{82.2} &84.0 &\uline{86.6} &\uline{86.9} &\textbf{87.7} &\textbf{88.3} \\ \hline
\multicolumn{1}{c|}{}    & w/o \textsf{S} 
&\textbf{40.5} &\textbf{42.4}  &\textbf{43.6}  &\textbf{44.2} &\textbf{44.3}  &\textbf{45.0}  &\textbf{45.2} \\
\multicolumn{1}{c|}{CIFAR-10}    & w/o \textsf{FAm} 
&\uline{40.0} &41.8  &42.7  &43.5 &\uline{44.2}  &\textbf{45.0}  &\uline{45.1} \\
\multicolumn{1}{c|}{}    & w \FAMS{} [ours] 
&\uline{40.0} &\uline{42.3}  &\uline{42.9}  &\uline{43.5} &44.1  &\uline{44.8}  &\textbf{45.2} \\
\bottomrule
\end{tabular}
}
\label{tab:effect_random}
\end{table}

\subsubsection{Effect of Randomness in \FAMS{}}
\label{ssec:exp_random}
To address concerns about the potential effect of \textbf{randomness} in subset sampling and the awakening process on the efficacy of CHASe, we conduct an analysis to gradually remove components of \FAMS{} and investigate the accuracy of the resulting approaches. Specifically, we examine the accuracy of approaches without the subset sampling (denoted as `w/o \textsf{S}') and the complete \FAMS{} (denoted as `w/o \textsf{FAm}').

As shown in \cref{tab:effect_random}, the complete \model{} maintains sub-optimal or even optimal accuracy when compared to approaches without the subset sampling and \FAMS{}. This finding suggests that the effect of \FAMS{} on the final accuracy is relatively minor and that the effectiveness of \model{} is mainly due to its ability to select informative samples using quantified EVs and the alignment loss term. Overall, our analysis demonstrates that \model{} is \emph{insensitive} to the randomness introduced by \FAMS{}.

\subsection{In-Depth Analysis of Epistemic Variation (RQ4)}
\label{ssec:exp_evstudy}

\subsubsection{EV's Relevance to Model Accuracy}
\label{ssec:ev_relevance}

We plot the average EVs of samples per client against their corresponding test accuracy on CIFAR10 under the \textsf{AbCo} setting\footnote{Here, we opt for \textsf{AbCo} over \textsf{ReCo} to purely analyze the heterogeneity inherent in the clients' data distributions.} in \cref{tab:ev_elim} to study whether they contribute to the model accuracy. Our results show an \emph{inverse correlation} between EV and model accuracy, suggesting that higher accuracy corresponds to lower average EVs of samples, across all baseline methods. Notably, the proposed \model{} not only achieves the fastest improvement in accuracy, but also significantly reduces EVs during training. It is important to note that different methods measure sample information from different dimensions (e.g., uncertainty, diversity, etc.), so the decrease in EV may not be directly proportional to the increase in accuracy for each method. Nonetheless, our findings support the effectiveness of EV in identifying important samples and suggest that it provides good guidance in data selection.

\subsubsection{Alternative Strategies for Quantifying EVs}
\label{sssec:exp_evquant}

We compare our strategy of quantifying EVs (see \cref{subsec:CEV}) with the two nuanced strategies below.

\noindent\textbf{(1) Adding Class Number}.
To determine the modified EV, we tally the number of unique classes present in the sequential inference results and add this number to the original EV. For instance, in the example inference sequence [`dog', `cat', `cat', `zebra', `cat'] depicted in \cref{fig:Tloc}, there are three different classes (`dog', `cat', and `zebra'), resulting in a modified EV of 6 (i.e., 3 + 3). 

\begin{figure}[t]
\centering
    \begin{tabular}{cc}
        \begin{minipage}[t]{1.6in}
        \includegraphics[width=1.85in]{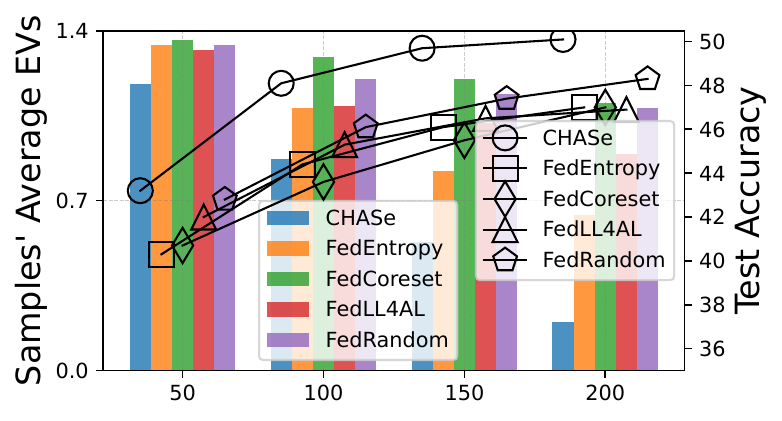}
        \vspace{-0.7cm}
        \caption{EV vs Accuracy.}
        \label{tab:ev_elim}
        \end{minipage}  
    \hspace{2.5mm}
        \begin{minipage}[t]{1.6in}
        \includegraphics[width=1.6in]{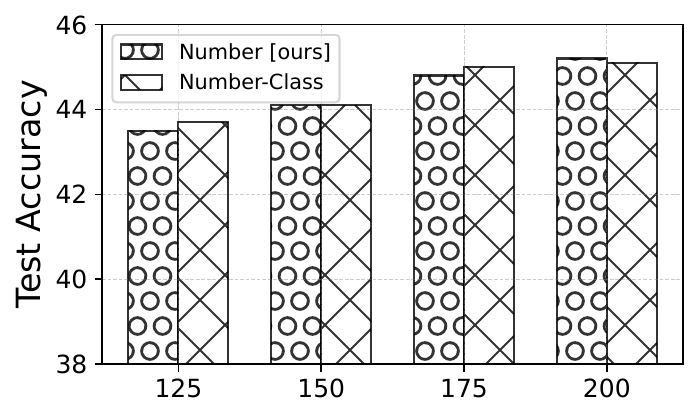}
        \vspace{-0.7cm}
        \caption{Class number.}
        \label{tab:class_number}
        \end{minipage}
    \end{tabular}
\end{figure}

We compare the accuracy of our original strategy (`Number') with the accuracy of the new strategy (`Number-Class') on CIFAR-10 under the \textsf{ReCo} setting, as presented in \cref{tab:class_number}. The accuracy differences between the two strategies are -0.2, 0.0, 0.2, and -0.1 at rounds 125, 150, 175 and 200, respectively. The improvement is \emph{not} significant, and we ultimately adopt the original strategy in \model{} due to its simplicity.

    

\noindent\textbf{(2) Additional Global EV (GEV).}
GEV is calculated as the absolute difference of the entropy before and after model aggregation at the global model. 
Specifically, after all updated local models $\boldsymbol{\omega}_{k}^{r}|_{k=1}^{K}$
are uploaded, the server aggregates them to obtain the new global model $\boldsymbol{\omega}^{r}$ and broadcasts it to all clients.
For client $k$, 
\model{} will re-infer on $\mathcal{D}_{k}^{U}$ with $\boldsymbol{\omega}^{r}$ and calculate the inference confidence $\mathcal{Q}(\boldsymbol{\omega}^{r};\boldsymbol{x}^{i})$:
\begin{align}
    \mathcal{Q}(\boldsymbol{\omega}^{r};\boldsymbol{x}^{i}) &=-\sum_{c=1}^{C} p(y_{c}^{i} \mid  \boldsymbol{\omega}^{r} ;\boldsymbol{x}^{i}) \log p(y_{c}^{i} \mid  \boldsymbol{\omega}^{r} ;\boldsymbol{x}^{i}).
    \label{entropy}
\end{align}

Then, \model{} calculates the inference confidence variation ${GEV}_{k}^{r}$ before and after model aggregation by the absolute difference of the entropy as follows:
\begin{align}
    &{GEV}_{k}^{r} =|\mathcal{Q}(\boldsymbol{\omega}^{r};\boldsymbol{x}^{i})-\mathcal{Q}(\boldsymbol{\omega}_{k}^{r};\boldsymbol{x}^{i})|.
    \label{delta_glo}
\end{align}

Finally, our acquisition function is defined as follows:
\begin{equation}
\mathcal{D}_{k}^{S} = \{ \boldsymbol{x}^{i}| A({EV}_{k}^{r}(\boldsymbol{x}^{i})+{GEV}_{k}^{r}(\boldsymbol{x}^{i});\mathcal{B}_{k})\}.
\end{equation}

As shown in \cref{tab:ablation_gev}, the impact of GEV is found to be limited. In \fram{}, even without global information, we can effectively quantify EVs because the server distributes the aggregated model to clients in each round, and each client trains locally on the received global model. 
This way, EVs generated by the server's aggregation on heterogeneous clients have been directly fed back to the client's local training. By quantifying EVs locally, we sample data that are vulnerable to interference from other clients, and this helps alleviate the oscillation of the decision boundary caused by heterogeneous clients. This is the fundamental difference between \fram{} and traditional AL.
Thus, additional GEVs are not employed.
\begin{figure}[t]
\centering
\includegraphics[width=0.48\textwidth]{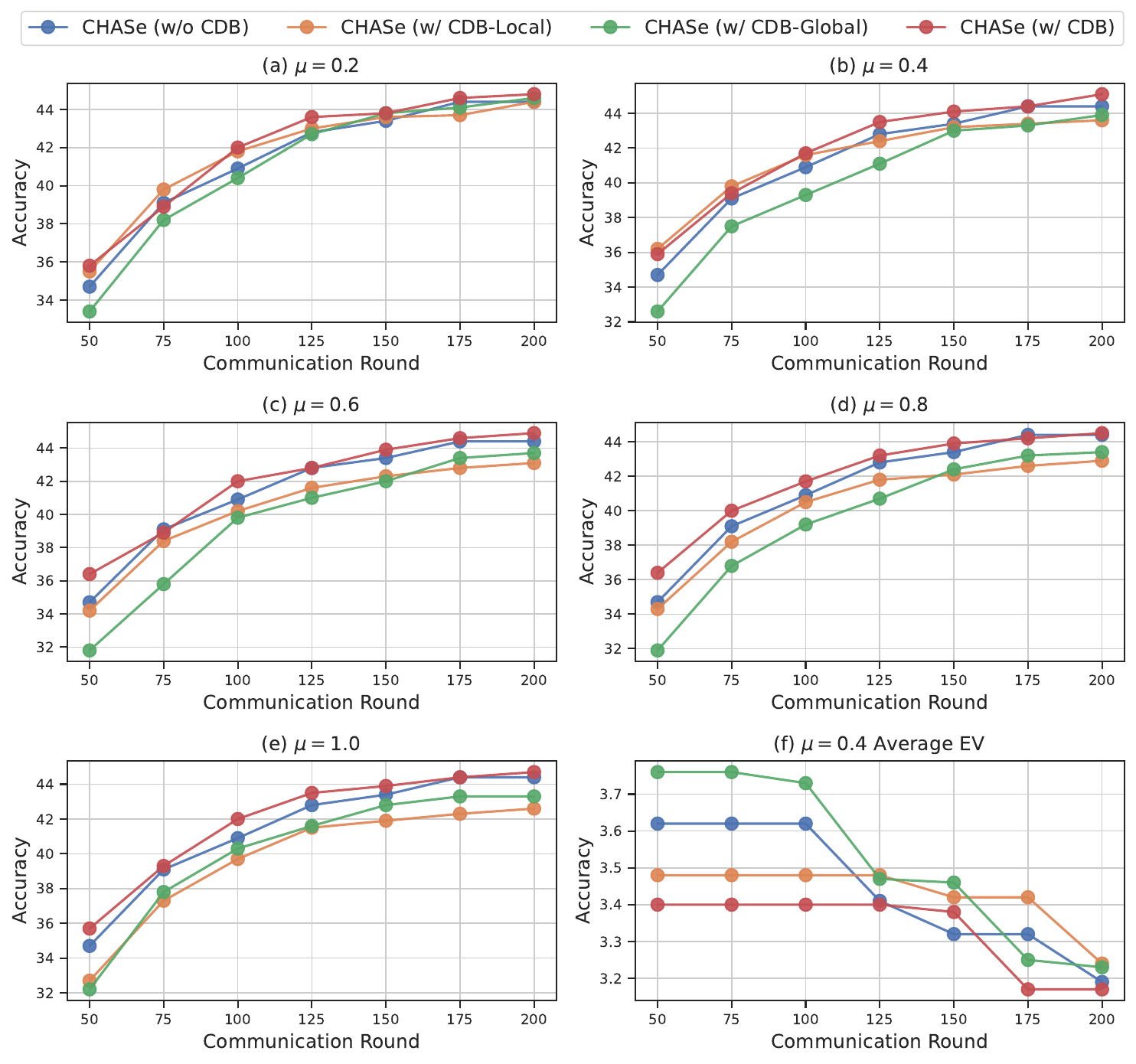}
\vspace{-0.3cm}
\caption{Ablation study of alignment loss terms for decision boundary calibration in the CIFAR-10 and \textsf{ReCo} scenario.}
\label{fig:abla_cdb}
\end{figure}

\begin{table}[!htbp]
     \caption{
        Local EV vs global EV under \textsf{ReCo}.
    }
  \centering
    \footnotesize
    \renewcommand{\arraystretch}{0.6}
    \setlength{\tabcolsep}{1.5mm}{
    \begin{tabular}{ccc|ccccccc}
    \toprule
    \multicolumn{3}{c|}{\model{}}  & \multicolumn{7}{c}{Test Accuracy (\%) at Round $r$} \\
    \midrule
    \multicolumn{1}{c|}{Datasets} &EV & GEV     & 50    & 75    & 100   & 125   & 150   & 175  & 200 \\
    \midrule

    \multicolumn{1}{c|}{\multirow{3}{*}{CIFAR-10}} & &  
    &39.2 &40.7 &41.6 &42.3 &43.1 &43.7 &43.9 \\ 
    
    \multicolumn{1}{c|}{}&+ &           
    &\textbf{39.8} &41.3 &42.6 &\textbf{43.6} &43.7 &\textbf{44.8} &\textbf{44.8} \\

    \multicolumn{1}{c|}{}&+ &+ 
    &39.7 &\textbf{41.9} &\textbf{42.9} &43.3 &\textbf{44.3} &44.4 &\textbf{44.8} \\
    \midrule
    \multicolumn{1}{c|}{\multirow{3}{*}{CIFAR-100}} & &  
    &20.7 &23.6 &25.1 &26.6 &27.4 &28.2 &29.0 \\ 
    
    \multicolumn{1}{c|}{}&+ &           
    &\textbf{21.4} &\textbf{24.1} &\textbf{26.1} &27.1 &\textbf{28.2} &\textbf{29.2} &\textbf{30.0}  \\

    \multicolumn{1}{c|}{}&+ &+ 
    &21.3 &24.0 &26.0 &\textbf{27.2} &28.1 &\textbf{29.2} &29.9 \\
   
    \bottomrule
    \end{tabular}
    }
  \label{tab:ablation_gev}
\end{table}

\subsection{Exploration of Ablation, Baseline \& Setting}
\label{ssec:exp_rebuttal}

\subsubsection{Ablation on Alignment Loss Terms of CDB}
\label{sssec:exp_cdb}

To validate the importance of aligning low-EV samples, we conduct experiments in the \textsf{ReCo} scenario, focusing on the alignment loss in the decision boundary calibration (CDB) of \model{}. We compare four baselines: 1) \model{} \textsf{(w/o CDB)}, where samples are selected based on EVs without alignment loss; 2) \model{} \textsf{(w/ CDB-Local)}, where low-EV samples are aligned with the previous local model; 3) \model{} \textsf{(w/ CDB-Global)}, where high-EV samples are aligned with the previous global model; and 4) \model{} \textsf{(w/ CDB)}, where low-EV samples are aligned to the previous local model and high-EV samples to the previous global model.

To avoid introducing extra noise, we run experiments three times with different random seeds on the CIFAR-10 dataset without data augmentation. The experiments varied the tradeoff loss parameters ($u=0.2, 0.4, 0.6, 0.8$), and the averaged results are presented in~\text{\cref{fig:abla_cdb}}.

First, \text{\cref{fig:abla_cdb}}(a$\sim$e) 
show that across varying $\mu$ values from $0.2$ to $0.8$, \model{} \textsf{(w/ CDB)} consistently outperforms the baselines. This highlights the effectiveness of our CDB technique, which classifies samples based on EV and aligns their embeddings with more accurate models. This technique works synergistically to reduce the selection of noisy samples, ultimately improving global model performance.

Furthermore, aligning only low-EV or high-EV samples tends to degrade overall performance, at times performing worse than \model{} \textsf{(w/o CDB)}. This degradation can be attributed to the diluted contribution to the classification loss during local training. As the tradeoff 
$\mu$ increases from $0.4$ to $0.8$, this negative effect becomes more pronounced, whereas at smaller values ($\mu = 0.2$), the performance gap between \model{} variants remains relatively small.

To enhance interpretability, we analyze the impact of neglecting the alignment of low-EV samples, which can falsely inflate their EVs and lead to the selection of less informative samples. We track the minimum average EV of clients' unlabeled data at $\mu = 0.4$, as shown in ~\text{\cref{fig:abla_cdb}}(f). 
Notably, there is an inverse relationship between average EV and performance. 
\model{} \textsf{(w/ CDB)}- exhibits the greatest reduction in average EV, and also achieves the highest performance. 
Conversely, other baselines show smaller EV reductions and poorer performance. 
A notable turning point is at round 150, where \model{} \textsf{(w/ CDB-Global)} lags behind \textsf{(w/ CDB-Local)} due to a higher average EV. However, \model{} \textsf{(w/ CDB-Global)} improves and surpasses \textsf{(w/ CDB-Local)} between rounds 175 and 200 as its average EV decreases. This inverse correlation between EV reduction and performance confirms the critical role of EV in enhancing FAL performance.

\begin{figure}[t]
\centering
\begin{minipage}{0.5\textwidth}
    \centering
    \includegraphics[width=0.7\textwidth]{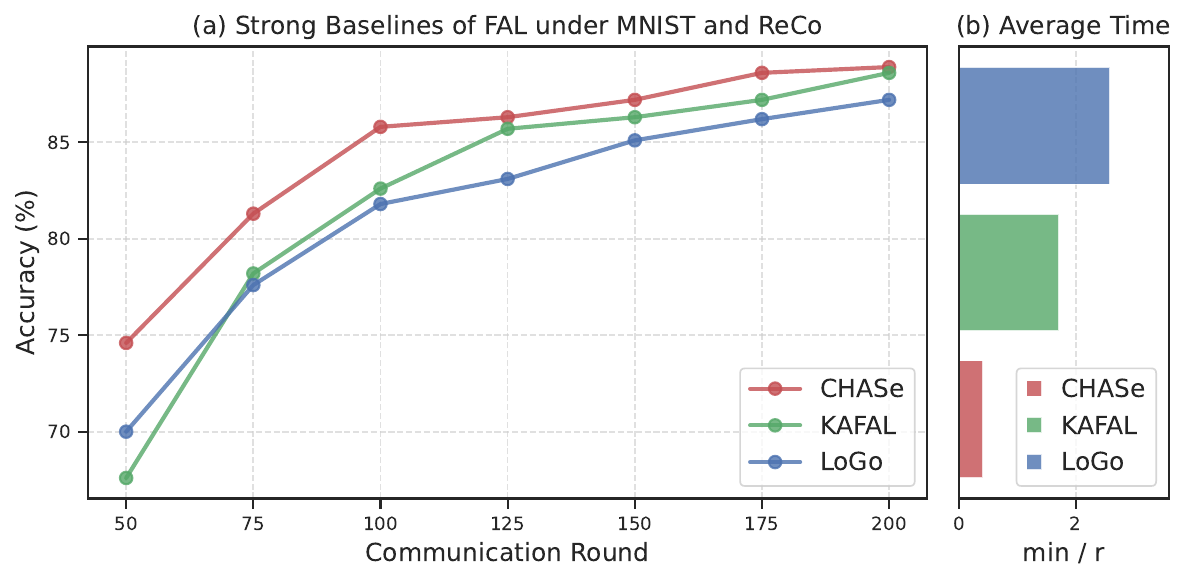}
    \caption{Performance comparison of \model{} with additional strong baselines, KAFAL\cite{cao2022knowledgeaware} and LoGo\cite{kim2023rethinking}, evaluated on the MNIST under the \textsf{ReCo} scenario.}
    \label{fig:abla_strong_baselines_mnist}
\end{minipage}  
\centering
\begin{minipage}{0.5\textwidth}
    \centering
    \includegraphics[width=0.7\textwidth]{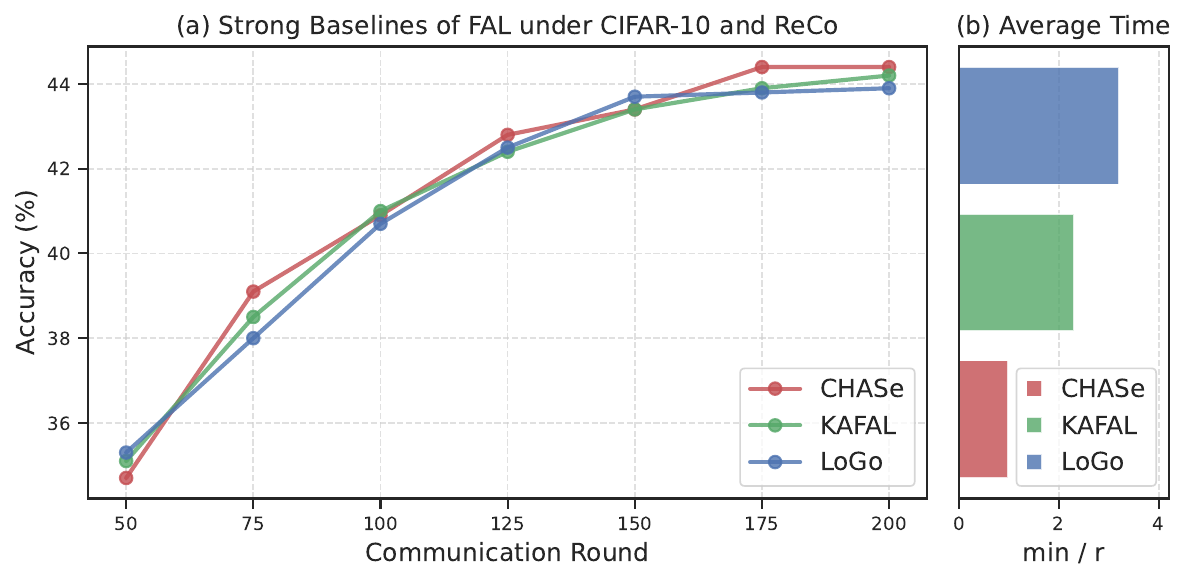}
    \caption{Performance comparison of \model{} with additional strong baselines, KAFAL\cite{cao2022knowledgeaware} and LoGo\cite{kim2023rethinking}, evaluated on the CIFAR-10 under the \textsf{ReCo} scenario.}
    \label{fig:abla_strong_baselines_cifar}
\end{minipage}  
\end{figure}
\subsubsection{Comparison on Strong Baselines}
To ensure a comprehensive performance comparison, we added strong baselines that use both local and global models to select data, namely KAFAL \text{\cite{cao2022knowledgeaware}} and LoGo \text{\cite{kim2023rethinking}}. 
Both our method and the baselines adhere strictly to the FedAvg process without introducing additional loss terms. 
For LoGo, we utilize the locally trained model from the previous round to extract features from the unlabeled data, and then cluster the feature embeddings using K-Means. 
The evaluation is conducted on the unaugmented MNIST and CIFAR-10 datasets under the challenging \textsf{ReCo} scenario, with averaged results from three experiments shown in \text{\cref{fig:abla_strong_baselines_mnist,fig:abla_strong_baselines_cifar}}.

\text{\cref{fig:abla_strong_baselines_mnist}}(a) and \text{\cref{fig:abla_strong_baselines_cifar}}(a) illustrate that our approach consistently outperforms KAFAL and LoGo in terms of global model performance. This underscores the effectiveness of our EV-based data selection method, which evaluates multiple decision boundaries to reduce noisy sample selection, thereby enhancing global model performance. In contrast, KAFAL and LoGo rely on inaccurate instant global and local models, which leads to the selection of lower-informative data and a limited performance gain.

Additionally, \text{\cref{fig:abla_strong_baselines_mnist}}(b) and \text{\cref{fig:abla_strong_baselines_cifar}}(b) show that \model{} significantly reduces the average execution time compared to KAFAL and LoGo. This efficiency is attributed to the \textsf{FAmS} technique, which substantially reduces \model{}'s computational overhead (as shown in \text{\cref{fig:SV}}). In comparison, LoGo's reliance on K-Means introduces significant complexity, increasing the overall time cost.

\begin{figure*}[t]
\centering
    \centering
    \includegraphics[width=0.9\textwidth]{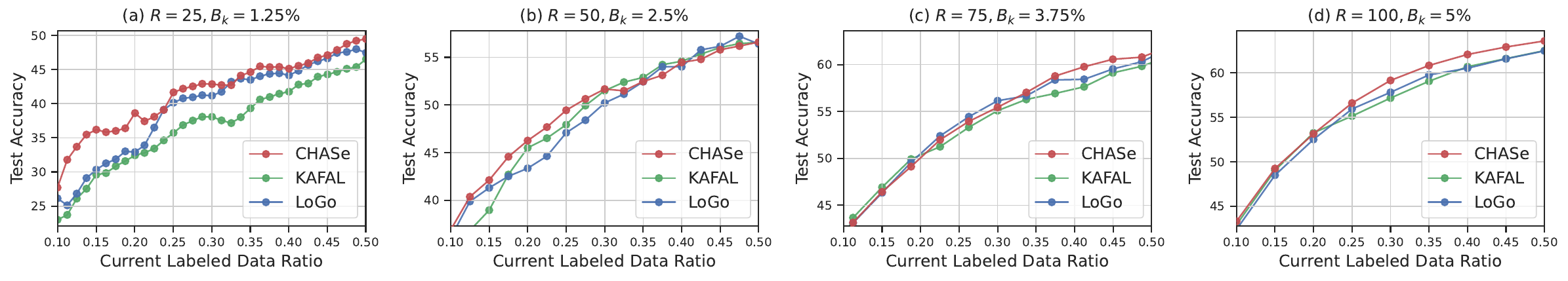}
    \caption{Adaption to a continual FAL setting with an adaptive annotation budget (simulated human annotators). The results are smoothed using a moving average with a window size of 3 to reduce noisy data.}
    \label{fig:abla_continual_FAL_budget}
\end{figure*}



\subsubsection{Adapting \model{} to FAL Using Random Reset}
\label{exp:continual_fal}
{In typical FL tasks like next-word prediction~\text{\cite{mcmahan2017communication}}, users often drop out after a round, and future participation is uncertain\text{\cite{saha2022colrel}}. Our FAL setting addresses this by ensuring that data selection occurs immediately when users join a communication round, maximizing valuable data collection even if their participation is brief.
Fortunately, \model{}, as a specific case of traditional FAL, can seamlessly adapt to the default setup of KAFAL\mbox{~\cite{cao2022knowledgeaware}} and LoGo\mbox{~\cite{kim2023rethinking}}. 
This adaptation is achieved by increasing the number of training rounds between sampling intervals while employing random initialization of the model before FL training.}

To explore the performance of \model{} in a traditional FAL setting, we increase the number of FL rounds to 25, 50, 75, and 100 and conduct experiments on CIFAR-10 ($C_k$=5), comparing against KAFAL and LoGo. To ensure fairness, no additional auxiliary losses are used. 
To improve inference efficiency, like LoGo and KAFAL, \model{}’s data selection relies only on the final FL communication round of the current sampling cycle.
To focus on realistic FAL, 
we simulate two scenarios: adaptive annotation and fixed annotation budget (see
supplementary). All other settings remain consistent with the previous CIFAR-10 experiments.

For the adaptive annotation budget scenario, clients participating in more FL rounds receive a larger budget per sampling round. This simulates scenarios where more active clients (akin human annotators) are allocated additional resources. For $R=25/50/75/100$, their annotation budget per sampling is $B_k=1.25\%/2.5\%/3.75\%/5\%$ of the original unlabeled data pool.
\text{\cref{fig:abla_continual_FAL_budget}} shows that \model{} consistently outperforms KAFAL and LoGo under varying round. In particular, \text{\cref{fig:abla_continual_FAL_budget}}(a) shows that with fewer FL training rounds and a smaller annotation budget, the model’s knowledge is relatively limited. Existing methods, which rely on selecting data using potentially inaccurate models, often result in suboptimal choices. In contrast, \model{} selects samples based on their high frequency across multiple decision boundaries, effectively mitigating the errors of single decision boundaries and reducing the inclusion of low-informative samples.

\section{Related Work}
\label{sec:related}


\noindent\textbf{Active Learning (AL)}.
AL has been widely applied to various tasks, including error detection \cite{ErroDetection}, image captioning \cite{ImageCaption}, and person re-identification \cite{PersonIdentification}.
AL methods can be categorized into three groups, namely uncertainty-based, distribution-based, and loss-based.

\emph{1) Uncertainty-based methods}: Uncertainty-based methods select data samples according to uncertainty,
which can be measured by the predicted probability \cite{lewis1994heterogeneous,lewis1994sequential},
the margin between the probability of two classes \cite{joshi2009multi},
or the entropy of prediction \cite{luo2013latent,settles2012active}. 
An intuitive assumption is that unlabeled data with high uncertainty might be beneficial for training better models.
In addition, Bayesian neural networks \cite{blundell2015weight}, MC-Dropout \cite{gal2016dropout}, or Ensembles are used to estimate model uncertainty.
Yet, such methods are computationally inefficient.
Notably, for AL-tailored temporal uncertainty methods like TOD\text{~\cite{huang2021semi}}, which measure uncertainty by comparing logits between two updated models, the main challenge in heterogeneous FAL is selecting which two models to compare.

\emph{2) Distribution-based methods}: This family of methods estimates the distribution of unlabeled samples to select important ones.
While discrete optimization methods \cite{guo2010active,yang2015multi} are used for a larger sample subset by considering diversity, the goal of the clustering method \cite{nguyen2004active} is to find the cluster centroids of subsets. 
The Core-set \cite{sener2018active} defines AL as a core-set selection problem, i.e.,~finding a small subset such that a model learned on them is competitive on the whole set, which addresses the failure of 
many AL heuristics when applied in a batch setting.

\emph{3) Loss-based methods}: These methods introduce additional networks to select hard samples by designing various loss functions.
An early learning loss approach, LL4AL \cite{yoo2019learning}, estimates samples' uncertainty and diversity and selects them with significant ``loss''. 
Recent works \cite{yuan2021multiple,fu2021agreement} progressively align data distributions for effective AL by introducing multiple adversarial classifiers with iterative loss optimization. 

The leading methods from these groups, including entropy-based \cite{luo2013latent}, Core-set \cite{sener2018active} and LL4AL \cite{yoo2019learning} have not explored the federation
, let alone the client heterogeneity.


\smallskip
\noindent\textbf{Federated Learning (FL) and Federated AL}.
FL was early discussed in literature such as study \cite{konevcny2015federated} and further developed along with the proposal of FedAvg \cite{mcmahan2017communication}.
Ahemd et al. \cite{ahmed2020active} and Mohammad et al. \cite{mohammad21flare} explore the effectiveness of current AL methods in FL, but they treat AL and FL as two completely orthogonal modules.
Jia et al. \cite{jia2019active} and Nicolas et al. \cite{nicolas2020combine} integrate the pipeline of AL into FL to reduce clients' training costs. 
Jin et al. \cite{jin2022federated} further verify that picking data on the global model (i.e., FAL($\bullet$)) performs better than that on the local model and the random sampling.
These above works show that existing AL methods perform decently in FL with IID.
However, these works ignore the FL's Non-IID issue \cite{10184650}.
Jia et al. \cite{jia2020Robust} discuss the resistance of local training epochs and aggregation frequency to heterogeneous data. However, it is still limited to the application of existing AL methods.
To mitigate the Non-IID issue in \fram{}, recent studies examine both local and global models to select samples with high intra- and inter-class diversity \text{\cite{cao2022knowledgeaware}} or samples with labels in majority \text{\cite{kim2023rethinking}}.
However, their selections rely on the instant models, which could vary significantly during the training, as disclosed by the samples' epistemic variation (see \text{\cref{fig:intro}}).
In contrast, the proposed \model{} approach considers the historical model states by capturing the epistemic variation for effective and robust data selection.

Besides, the notation of \emph{Active Federated Learning} (AFL) \cite{goetz2019active} clearly differs from \fram{} that we studied. AFL focuses on actively selecting clients from a server perspective in FL. 
There are also studies \cite{li2021sample,shin2022sample} on data selection in FL, but they assume data is fully labeled and do not involve AL.

\section{Conclusion}
\label{sec:conclusion}
This paper addresses the problem of heterogeneous federated active learning (FAL) for annotation data selection and model training while preserving data privacy. Existing FAL methods fall short in addressing client heterogeneity and variations in global and local model parameters during training. 
To address these issues, the proposed \model{} selects informative unlabeled samples with high epistemic variations (EVs), which reveal a sample's difficult level of being inferred by the local model. Various techniques are proposed to ensure the effectiveness and efficiency of \model{}.
The experimental results demonstrate improved FAL effectiveness across different types of datasets and models and heterogeneous scenarios.



\bibliographystyle{unsrt}
\bibliography{sample-base.bib}
\clearpage
\section*{\large Supplementary Material to the Paper Entitled ``\model{}: Client Heterogeneity-Aware Data Selection for Effective Federated Active Learning''}

\bigskip

\subsection{Key Notations}
The key notation in this paper is provided in \cref{tab:notation}.

\begin{table}[!htbp]
    \centering{
    \caption{Key notations.} \label{tab:notation}
    \setlength{\tabcolsep}{0.3mm}
    \begin{tabular}{@{}c|c@{}}
    \toprule
        Symbol &  Meaning \\
    \midrule
         $\boldsymbol{x}_i,y_i$ & The $i$-th sample and its corresponding truth label \\
         $\hat{y}_i$ & The predicted label of sample $x_i$\\
         $R$, $r$ & Overall communication round and the $r$-th round \\
         $E$, $e$ & Overall local training epoch and the $e$-th epoch \\
         $K$, $k$ & Overall size of clients and the client with serial number $k$ \\
         $\mathcal{B}$, $\mathcal{B}_{k}$ & Overall fixed budget and the budget of client $k$  \\
         $\mathcal{D}_{k}^{L}$, $\mathcal{D}_{k}^{U}$, $\mathcal{D}_{k}^{S}$ &
         Labeled, unlabeled, and selected sets at client $k$  \\
         $N_{k}^{L}$, $N_{k}^{U}$, $N_{k}^{S}$ &
         \emph{Size of} labeled, unlabeled, and selected sets at client $k$  \\
         $\tilde{\mathcal{D}}_{k}^{U}$ & Unlabeled subset of client $k$ by subset sampling \\
         $\mathcal{D}_{k}^{D}$ & Dormant set of client $k$\\
         $\mathcal{D}_{k}^{U_0}$ & Unlabeled data of client $k$ with zero EVs\\    
         $\boldsymbol{\omega}^{r-1}$ & The received global model of clients at round $r$\\
         $\boldsymbol{\omega}_{k}^{r}$ & The trained local model of client $k$ after $E$ epochs \\        
         $\tilde{\boldsymbol{\omega}}_{k}^{e}$ & The partially trained local model of client $k$ after $e$ epochs\\
         $\boldsymbol{V}_{k}^{r}(\boldsymbol{x}^{i})$ &  The historical variations on sample $\boldsymbol{x}_i$ of client $k$ at round $r$ \\
         ${EV}_{k}^{r}(\boldsymbol{x}^{i})$ & The epistemic variation on sample $\boldsymbol{x}_i$ of client $k$ at round $r$ \\
    \bottomrule
    \end{tabular}}
\end{table}

\subsection{Details of the Experiment Settings}
\label{ssec:settings_sup}

\subsubsection{Datasets}
\label{ssec:data}
We evaluate the efficacy of \model{} using well-known \textbf{image datasets} following studies \cite{9835327,9835537}:
\begin{itemize}[leftmargin=*]
\item MNIST: This dataset for digit recognition includes 60,000 training images and 10,000 test images.

\item EMNIST: As an extension, EMNIST contains 814,255 images (731,668 for training and 82,587 for testing), featuring numbers, upper and lower case letters across 62 categories.

\item CIFAR-10: This set comprises 60,000 32x32x3 pixel color images within 10 classes, split into 50,000 for training and 10,000 for testing.

\item CIFAR-100: Similar to CIFAR-10 but with 100 classes. Each class has 600 images, divided into 500 for training and 100 for testing.
\end{itemize}

CIFAR-100 and EMNIST are notable for their high number of classes, with EMNIST also being large-scale, providing varied environments for testing effectiveness and robustness.

To test \model{}'s generalizability, we also include the Shakespeare \textbf{text dataset} \cite{shakespeare}, which includes 4,226,15 samples across 80 categories, based on the complete works of William Shakespeare. Following Leaf's guidelines \cite{caldas2019leaf}, 90\% of these samples are used for training, with the remaining 10\% for testing in the next character prediction tasks.

\subsubsection{Baselines} 
\label{ssec:baseline}

We evaluate the performance of our method against the following baselines:
\begin{itemize}[leftmargin=*]
\item \textbf{FedRandom} that combines the FedAvg \cite{mcmahan2017communication} approach with the random sample selection for annotation.

\item \textbf{FedEntropy} that combines FedAvg with the entropy-based AL \cite{luo2013latent} that selects samples with high prediction entropy.

\item \textbf{FedCoreset} that combines FedAvg with the Core-set \cite{sener2018active} based AL.

\item \textbf{FedLL4AL} that combines FedAvg with LL4AL \cite{yoo2019learning}.
As an adaption, the additional loss prediction employed in LL4AL is updated and aggregated also following FedAvg.

\item \textbf{\fram{}($\bullet$)} that applies each classical AL method (i.e.,~Entropy, Core-set and LL4AL) to the \fram{} scheme \cite{jin2022federated} where data are being selected using the global model.

\end{itemize}

\subsubsection{Metrics}
\label{sssec:metrics}

We measure the \textbf{effectiveness} of all methods in terms of the accuracy on the testing set with a fixed number of communication rounds, aligning with prior AL works \cite{yoo2019learning, fu2021agreement}.
We report the average accuracy measures on 3 runs with different seeds.
On the other hand, we measure the \textbf{efficiency} of all methods in terms of both execution time and inference cost per \fram{} round. Particularly, the inference cost refers to the number of inferred unlabeled samples.
We focus on reporting efficiency-related evaluation results in~\cref{sssec:exp_efficiency}.

\subsubsection{FL-related Setting}
\label{sssec:fl_setting}

We assume $K=10$ clients in MNIST, CIFAR-10 and CIFAR-100; and $K=100$ in the larger EMNIST and Shakespeare datasets.
We set the local epoch number $E=10$ and communication round $R=200$.
Following a previous work \cite{mcmahan2017communication} on simulating \emph{various degrees} of Non-IID settings, we specify each client has $C_k=$ 5, 2, and 10 classes for both labeled and unlabeled sets on CIFAR-10, MNIST and EMNIST, respectively.
Following another study \cite{Mikhail2019Bayesian} that specifically simulates the Non-IID setting for CIFAR-100, we use the Dirichlet distribution $Dir(\alpha)$ to generate heterogeneous data partitions with the concentration parameter $\alpha = 1e-2$.
For Shakespeare, we follow previous studies \cite{caldas2019leaf,mcmahan2017communication}, creating a Non-IID dataset with 1,146 clients representing each speaking role in each play. However, we only include clients with more than 10,000 samples and randomly select 100 of these clients to participate in \fram{}.

\subsubsection{AL-related Setting}
\label{sssec:al_setting}

To initiate the labeled set, we randomly select small data portions from the training set, specifically 4\% for CIFAR-10 and Shakespeare, 10\% for CIFAR-100, 1.33\% for MNIST, and 2\% for EMNIST. 
Regarding the annotation budget, we base our assumption on each client being honest and non-adversarial \cite{ipeirotis2014repeated}, thus not providing error labels.
We examine two distinct client behavior scenarios: 
\begin{itemize}[leftmargin=*]
\item \textsf{AbCo} [\textsf{Absolute Cooperation}]: All clients are fully engaged in the annotation. 
The requirement per client per round is $\mathcal{B}_{k}=10$ samples, e.g., given $K=10$ clients, the server's total budget per round is $\mathcal{B}= \mathcal{B}_{k} \times K = 100$.

\item \textsf{ReCo} [\textsf{Relative Cooperation}]: Not all clients are equally cooperative.
We use the Gaussian distribution to organize clients into passive, ordinary, and aggressive groups, and their ratio is 2:6:2.
A passive/ordinary/aggressive client will annotate $\mathcal{B}_{k}=$ 5/7/10 samples every 5/3/1 round(s).

\end{itemize}

\subsubsection{Other Settings}
\label{sssec:other_setting}

Considering the limited storage space and computational resources of the clients, as in existing FL works, CNN (2 Conv + 2 FC) \cite{xie2019dba} is used for MNIST, EMNIST and CIFAR-10, and VGG-19 for CIFAR-100.
We use the SGD optimization with a learning rate of 1e-3 for MNIST, EMNIST, and CIFAR-10, and 1e-2 for CIFAR100. 
For Shakespeare, following \cite{mcmahan2017communication}, we train two layers stacked character-level LSTM with a learning rate of 1.47 to predict the next character.
For the alignment loss $\ell_\text{align}$, we set $\tau = 0.5$ following the work \cite{Qinbin2021Contra}, $\mu =$ 0.1, 0.1, 1, 1, 0.3 for MNIST, EMNIST, CIFAR-10, CIFAR-100, and Shakespeare respectively.
For \FAMS{}, we set the subset size $N_s^U = 500$, the awaken threshold $\mathcal{B}'=3 \times \mathcal{B}_k$ and the awaken ratio $\beta=0.4$.
The experiments on MNIST, CIFAR-10, and Shakespeare are performed on a single machine with Quadro RTX 6000 GPUs and PyTorch 1.6.0, while those on EMNIST and CIFAR-100 use Quadro RTX A5000 GPUs and PyTorch 1.8.0.
The detailed hyperparameters for training are listed in~\cref{tab:hyperparameter}.

\begin{table}[!htbp]
    \caption{The hyperparameters associated with datasets.}
    \centering
    \footnotesize
    \renewcommand{\arraystretch}{1.0}
    \setlength{\tabcolsep}{0.2mm}{
    \begin{tabular}{l|ccccc}
    \toprule
        Hyperparameter & MNIST & EMNIST  &CIFAR-10  & CIFAR-100 & Shakespeare\\
    \midrule
         Learning Rate & 0.001 & 0.001 & 0.001 & 0.01 & 1.47\\
         Dropout Probability & 0.5 & 0.5 & 0.5 & 0.0 & 0.2 \\
         Rounds ($R$) & 200 & 200 &200 &200 &200\\
         Local Batch Size($B$) & 10 & 10 & 10 & 10 & 10\\
         Local Epoch($E$) & 10 & 10 & 10 & 10 & 10\\
         Size of Clients($K$) & 10 & 100 & 10 & 10 & 100\\
         Subset Size($N_S^U$) &500 &500 &500 &500 &500\\
         Awaken Ratio ($\beta$) &0.4 &0.4 &0.4 &0.4 &0.4\\
         Temperature ($\tau$) &0.5 &0.5 &0.5 &0.5 &0.5\\
         Weight in $\ell_\text{align}$ ($\mu$) &0.1 &0.1 &1 &1 &0.3\\
    \bottomrule
    \end{tabular}}
    \label{tab:hyperparameter}
\end{table}

\begin{figure*}[t]
\centering

    \centering
    \includegraphics[width=0.96\textwidth]{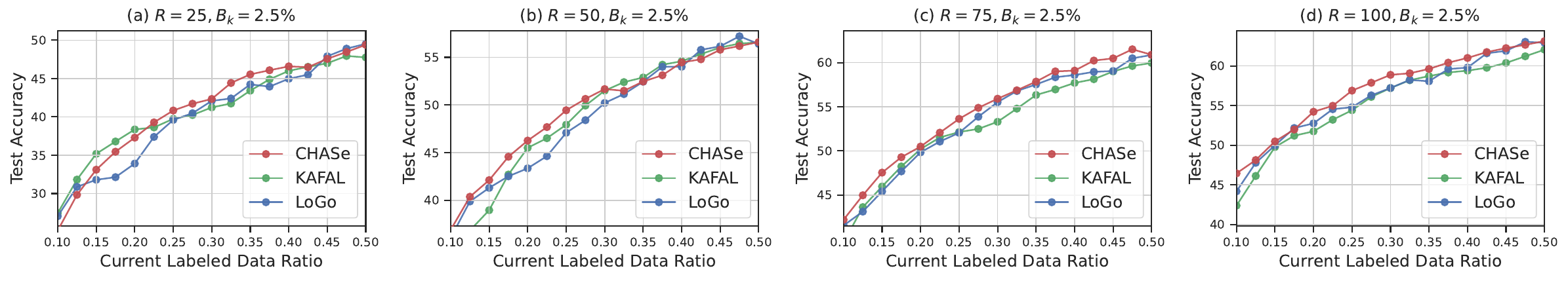}
    \caption{Adaption to a continual FAL setting with a fixed annotation budget (simulated machine annotators). The results are smoothed using a moving average with a window size of 3 to reduce noise data.}
    \label{fig:abla_continual_FAL_round}

\end{figure*}
\subsection{Adapting \model{} to FAL Using Random Reset}\label{exp:fixed_continual_fal}
Here, we explore the traditional FAL with a fixed annotation budget. In this scenario, the annotation capacity of clients remains limited and constant, regardless of the number of training rounds. This reflects cases where clients rely on automated or resource-constrained methods (e.g., machine annotators), allowing them to label a fixed amount of data ($B_k=2.5\%$) per sampling interval, irrespective of communication frequency.

\cref{fig:abla_continual_FAL_round} indicates that our method is always better or closer to KAFAL and LoGo. 
However, because the model itself has not learned enough knowledge in this scenario, the importance of unlabeled samples under the relatively large budget, therefore, the superiority of the strategy is easily covered under this tradition FAL.

\subsection{Visualization of Ablation Study}

\begin{figure}[!htbp] 
\centering
\subfigure[CIFAR-100 / Dir(1e-2) / \textsf{ReCo}]{ 
\begin{minipage}{0.5\linewidth}
\label{subfig:CIFAR100-GL-RCS-SAL_Abla}
    \includegraphics[
    width=4.2cm,
    ]{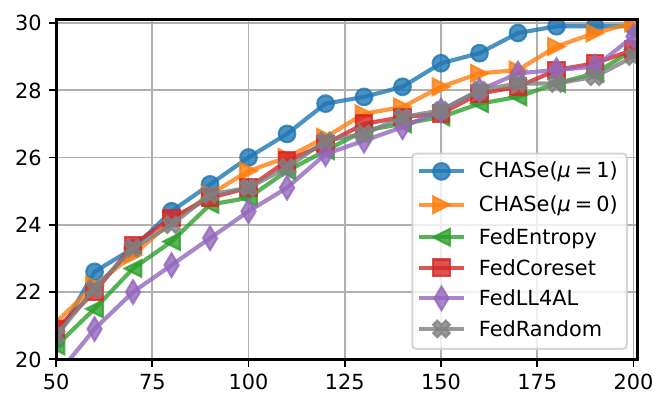}
\end{minipage}%
}%
\subfigure[EMNIST / $C_k$=10 / \textsf{ReCo}]{
\begin{minipage}{0.5\linewidth}
    \label{subfig:EMNIST-GL-RCS-SAL_Abla}
    \includegraphics[
    width=4.2cm,
    ]{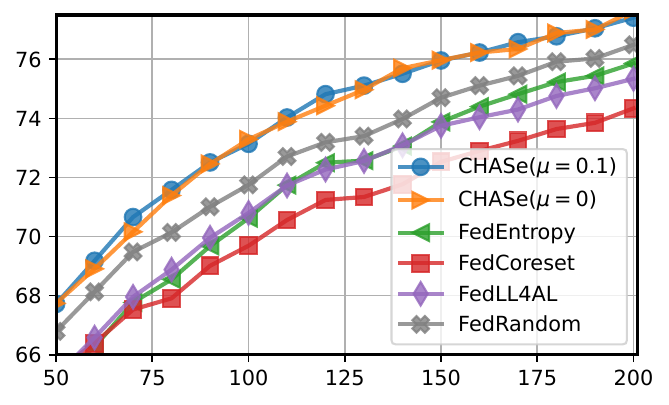}
\end{minipage}%
}%
\caption{Test accuracy curves on CIFAR-100 and EMNIST.}
\end{figure}

To ensure a fair comparison, we conducted additional studies by removing the alignment loss and evaluating \model{} without it, as illustrated in \cref{fig:text_exp} on Page 8 and \cref{tab:ablation} \& \cref{tab:P&C} on Page 9 of the manuscript. Space constraints in the original manuscript limited our discussion to CIFAR-10, but here we provide further assurance by presenting additional ablation results for CIFAR-100 and EMNIST, shown in \cref{subfig:CIFAR100-GL-RCS-SAL_Abla,subfig:EMNIST-GL-RCS-SAL_Abla} of this letter. 
These experiments demonstrate that \model{} ($u$ = 0) consistently outperforms other baselines, with decision boundary calibration further improving task performance under the same communication round.

\begin{figure}[!htbp]
    \centering
    \includegraphics[width=1\linewidth]{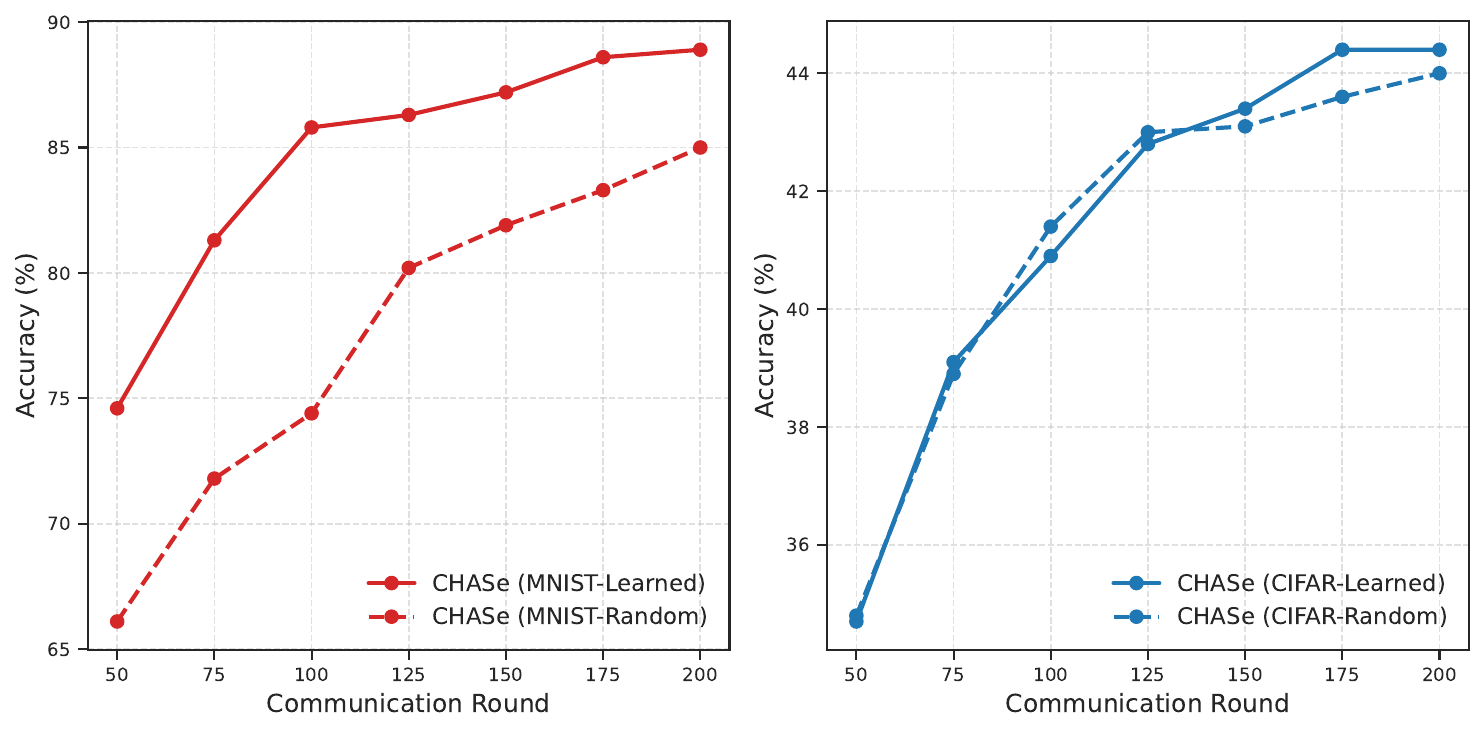}
    \caption{\textsf{CHASe} Performance on MNIST and CIFAR-10 under the \textsf{ReCo} scenario: Learned Model vs. Random Model.}
    \label{fig:abla_randset}
\end{figure}

\subsection{Learned Model vs. Random Model}

{We explore the effects of leveraging the knowledge learned from the current communication round versus using a randomly initialized model on \textsf{CHASe} performance in the \textsf{Reco} scenarios for both MNIST and CIFAR-10. Notably, we did not incorporate any auxiliary loss terms other than the classification loss. Specifically, as shown in~\text{\cref{fig:abla_randset}}, using the current model outperforms the random initialization model across different tasks. Moreover, as shown in the left side of~\text{\cref{fig:abla_randset}}, for simpler task (e.g.~MNIST) models, the learning process is less susceptible to interference, leading to greater benefits. The results indicate that leveraging the knowledge gained from the current communication round in our FAL setup is crucial for maximizing the labeling capability of clients with uncertain participation, such as those that may drop out due to network interruptions or tasks like next-word prediction where client engagement is intermittent, thereby improving model performance.}

\end{document}